\documentclass[journal]{IEEEtran}
\usepackage{times}
\usepackage{epsfig}
\usepackage{graphicx}
\usepackage{amsmath}
\usepackage{amssymb}
\usepackage{url}
\usepackage[dvipsnames]{xcolor}
\usepackage{hyperref}

\usepackage{threeparttable}

\usepackage[ruled,vlined]{algorithm2e}

\SetCommentSty{mycommfont}
\SetKwInput{KwInput}{Input}
\SetKwInput{KwOutput}{Output}
\SetKwInput{KwParameters}{Parameters}

\ifCLASSINFOpdf
\else
\fi
\begin{document}
%
\title{Sub-Goal Social Force Model for Collective Pedestrian Motion Under Vehicle Influence}
%
%
%

\author{Dongfang Yang,~\IEEEmembership{Student Member,~IEEE,}
        Fatema T. Johora,~
        Keith A. Redmill,~\IEEEmembership{Senior Member,~IEEE,}\\
        Ümit Özgüner,~\IEEEmembership{Life~Fellow,~IEEE,}
        and~Jörg P. Müller
\thanks{Dongfang Yang, Keith Redmill, and Ümit Özgüner are with the Department of Electrical and Computer Engineering, The Ohio State University, Columbus, OH 43212 USA.}
\thanks{Fatema T. Johora and Jörg P. Müller are with Department of Informatics, Clausthal University of Technology, Julius-Albert-Str. 4, 38678 Clausthal-Zellerfeld, Germany.}
\thanks{This work has been submitted to the IEEE for possible publication. Copyright may be transferred without notice, after which this version may no longer be accessible.}}

%
%

\markboth{Journal of \LaTeX\ Class Files,~Vol.~xx, No.~x, Month~Year}%
{Shell \MakeLowercase{\textit{et al.}}: Bare Demo of IEEEtran.cls for IEEE Journals}
%



\maketitle

\begin{abstract}
In mixed traffic scenarios, a certain number of pedestrians might coexist in a small area while interacting with vehicles. In this situation, every pedestrian must simultaneously react to the surrounding pedestrians and vehicles. Analytical modeling of such collective pedestrian motion can benefit intelligent transportation practices like shared space design and urban autonomous driving. This work proposed the sub-goal social force model (SG-SFM) to describe the collective pedestrian motion under vehicle influence. The proposed model introduced a new design of vehicle influence on pedestrian motion, which was smoothly combined with the influence of surrounding pedestrians using the sub-goal concept. This model aims to describe generalized pedestrian motion, i.e., it is applicable to various vehicle-pedestrian interaction patterns. The generalization was verified by both quantitative and qualitative evaluation. The quantitative evaluation was conducted to reproduce pedestrian motion in three different datasets, HBS, CITR, and DUT. It also compared two different ways of calibrating the model parameters. The qualitative evaluation examined the simulation of collective pedestrian motion in a series of fundamental vehicle-pedestrian interaction scenarios. The above evaluation results demonstrated the effectiveness of the proposed model.

\end{abstract}


%
\IEEEpeerreviewmaketitle


\section{Introduction}

\IEEEPARstart{F}{uture} transportation systems are expected to become a highly mixed mode full of different types of traffic participants. This is an inevitable trend due to the development of urban areas. Mixed traffic scenarios imply more frequent vehicle-pedestrian interaction. In some cases, a certain number of pedestrians might coexist in a small area while interacting with vehicles. This requires each pedestrian to simultaneously react to the surrounding pedestrians and vehicles. We call this phenomenon the collective pedestrian motion under vehicle influence. This work focuses on the analytical modeling of such pedestrian motion. 


The analytical modeling can benefit various intelligent transportation practices. For example, researchers can utilize the pedestrian model to simulate pedestrian behavior in mixed traffic scenarios. This could be analyzing the shared space design by generating collective pedestrian motion alongside moving vehicles in the shared space area.
For another example, the pedestrian model can be used to create pedestrian-involved scenarios for the testing of urban autonomous driving algorithms. The scenarios could be customized to be uncommon or even adversarial, which are difficult to be observed in the real world. Furthermore, the pedestrian model could synthesize different pedestrian motion data that can possibly help the training of some data-driven approaches for pedestrian behavior modeling. 

Within this context, we propose a generalized pedestrian model to describe the collective pedestrian motion under vehicle influence. A generalized model implies that the model is applicable to a variety of vehicle-pedestrian interaction scenarios. The scenario could be as simple as one pedestrian interacting with one vehicle or as complex as several groups of pedestrians interacting with a convoy of vehicles.


Early pedestrian models primarily focus on pure pedestrian-to-pedestrian interaction. In recent years, some pedestrian models started to consider vehicle-to-pedestrian interaction. However, these models have some drawbacks. Some models are by design not generalized enough~\cite{chen2019assessing, zeng2017specification, pascucci2017discrete}. Some do have a generalized design, but they are not comprehensively evaluated~\cite{pascucci2015modeling, anvari2015modelling, rinke2017multi, johora2018modeling, yang2020social}. To fill these gaps, we proposed the sub-goal social force model (SG-SFM), which is a generalized model, and systematically evaluated it to verify the generalization.

The proposed SG-SFM builds on the original social force model (SFM)~\cite{helbing1995social, helbing2000simulating}, which was proposed for microscopic crowd motion modeling. We selected the social force based modeling for two reasons. 
First, the SFM is a suitable and popular choice for describing collective pedestrian motion. It treats each pedestrian as a point mass of second-order dynamics that is subject to a conceptualized virtual force as the action due to the interaction effect. This makes it easy to be applied to any number of pedestrians at the same time. 
Second, the vehicle influence can be translated into an additional portion of the virtual force so that the ego pedestrian can simultaneously consider the surrounding pedestrians and vehicles. 
Following this reasoning, the SG-SFM introduced a new design of vehicle influence on the pedestrian motion. To integrate the vehicle influence as part of the virtual force, we utilized the sub-goal concept~\cite{tamura2012development} to combine vehicle-to-pedestrian interaction with pedestrian-to-pedestrian interaction. In the original SFM, the pedestrian is directly driven to the final destination while considering the pedestrian-to-pedestrian interaction. But in the SG-SFM, with the sub-goal concept, a temporary goal is selected based on the direction of the final destination and on how the ego pedestrian is interacting with surrounding pedestrians and vehicles. 
As a result, the virtual force is composed of two types of sub-forces. The first type describes the \textit{reactive} action as the result of pure repulsion from interacting agents. The second type describes the \textit{proactive} action that accounts for both the navigation to the final destination and the navigation around the interacting agents.

Once the pedestrian model is ready, the model parameters need to be carefully determined so that the model can achieve its best performance. This is usually done by repeating the process of manually tuning the parameters and visually inspecting the simulation results. The simulation refers to simulating the pedestrian motion under vehicle influence using the model. If a dataset is available, the parameters can be further refined by conducting parameter calibration. A common approach for calibration is to minimize the difference between the simulated pedestrian trajectory and the ground truth pedestrian trajectory. In this work, both manual tuning and calibration were applied. And the calibration employed the genetic algorithm. However, instead of just applying one dataset as was done in previous works~\cite{seer2014validating, dias2018calibrating, zeng2017specification}, a total of three different datasets~\cite{pascucci2017discrete, yang2019top} were applied. By showing good quantitative results over three different datasets, both the effectiveness and the generalization of SG-SFM was verified.

This work also explored calibrating the model parameters into different groups for different pedestrian characteristics. This is referred to as group calibration, while the common approach seeking only one group of parameters is referred to as universal calibration. The group calibration was done by firstly classifying pedestrians into different groups using K-means~\cite{marutho2018determination} and then calibrating a unique model for each group. 
Theoretically, the model refined by group calibration should have better performance than that by universal calibration. This was confirmed in the evaluation results, which serves as an additional clue to demonstrate the model effectiveness.

The pedestrian model can be evaluated in two ways.
The first one is to simulate a number of pedestrians in vehicle-pedestrian interaction scenarios and observe if the pedestrian motion is realistic or not. The second one uses datasets. The pedestrian model is evaluated by calculating the scores of quantitative metrics like average displacement error (ADE) and final displacement error (FDE). To comprehensively evaluate the proposed SG-SFM, both were applied in this work. The former is referred to simulation-based evaluation, while the latter data-based evaluation. A combined procedure of utilizing both evaluations was implemented. For the simulation-based evaluation, a series of systematically designed fundamental scenarios were proposed. They cover a variety of different types of vehicle influences on pedestrians. For the data-based evaluation, a better way of preparing data samples was adopted to ensure that the ego pedestrian trajectory in a data sample covers complete vehicle-pedestrian interaction. This subsequently led to a slight modification of the quantitative metrics. The new way of data sample preparation also applies to the parameter calibration.

The major contributions are summarized as follows:
\begin{itemize}
    \item A generalized model named sub-goal social force model (SG-SFM) was proposed to describe the collective pedestrian motion under vehicle influence. The model inherited the framework of the original social force model but was redesigned and optimized to incorporate the vehicle influence. The sub-goal design was adopted to combine the navigation to the final destination with the navigation around surrounding pedestrians and vehicles.
    \item The proposed model was calibrated and evaluated over three different datasets. A new way of creating data samples was employed to ensure that each sample contains complete vehicle-pedestrian interaction. Model parameter calibration explored two approaches, universal calibration and group calibration. Data-based evaluation over different datasets verified that the model is generalized enough to be applied in different scenarios. The results can be served as a benchmark for modeling the collective pedestrian motion under vehicle influence.
    \item Besides the data-based evaluation, this work also conducted simulation-based evaluation. This was achieved by running systematically designed fundamental scenarios of vehicle-pedestrian interaction. The evaluation results further demonstrated the effectiveness and the generalization of the model to a variety of interaction patterns.
    
\end{itemize}

In the rest of the paper, section~\ref{se:formulation} gives the problem formulation. Section~\ref{se:design_flow} describes the modeling process. The proposed SG-SFM is detailed in section~\ref{se:pedestrian_model}, followed by the model calibration in section~\ref{se:calibration} and the model evaluation in section~\ref{se:evaluation}. 
Sections~\ref{se:experiments} and~\ref{se:results} present the experiments and the corresponding results. Conclusions are drawn in section~\ref{se:conclusion}.

\section{Related Works}
\label{se:related}

\subsection{Pedestrian Motion Modeling}

\subsubsection{Analytical Modeling} 
Analytical pedestrian motion modeling could be either macroscopic or microscopic. Macroscopic modeling considers the motion of lots of pedestrians as a whole, in which properties such as pedestrian density and average walking speed and the change of them over time are the primary focus. Representative macroscopic pedestrian models are gas/fluid dynamics models~\cite{henderson1971statistics, helbing1992fluid}. Microscopic models consider the behavior of individual pedestrians and the interaction among them. Popular models include magnetic model~\cite{okazaki1979study}, cellular automata~\cite{toffoli1987cellular}, and social force model~\cite{helbing1995social, helbing2000simulating}. There are also some social force model variants~\cite{pellegrini2009you, yamaguchi2011you}, which basically follow a similar idea of how the interaction is designed in the original social force model. These models are primarily used for interactive pedestrian motion simulation. A comprehensive review can be found in~\cite{dong2019state}. 

\subsubsection{Data-driven Modeling}
With the advancement of machine learning and the availability of an increased amount of pedestrian data, data-driven modeling has attracted lots of attention in recent years. Data-driven models are slightly different from analytical models, as data-driven modeling is more preferable for tasks like prediction and estimation, but not for interactive pedestrian motion simulation. Although the pedestrian motion is still somehow considered in prediction and estimation, it is different from pure pedestrian motion modeling. Data-driven models usually require a context condition as the input, based on which the pedestrian's behavior such as motion and intention can be estimated and predicted. In Bayesian framework, dynamic Bayesian networks~\cite{kooij2019context} was successfully applied for pedestrian's intention and future motion prediction, conditioned on the current pedestrian state. Using deep neural networks~\cite{alahi2016social, sadeghian2019sophie}, future pedestrian trajectory considering interaction can be predicted based on historical pedestrian trajectory. One of the pioneer works is Social-LSTM~\cite{alahi2016social}. Reinforcement learning based approach~\cite{martinez2015strategies} is another way of modeling pedestrian motion, which is more close to pure pedestrian motion modeling while it still somehow utilizes the (synthesized) data. There is a bunch of literature focusing on data-driven approaches. Since data-driven modeling is not the main focus of this research, interested readers are recommended to see a survey~\cite{rudenko2020human} of pedestrian models.

\subsection{Social Force Model}

\subsubsection{Historical Development}
Social force model is one of the most popular methods for interactive pedestrian motion modeling. It was initially introduced in 1995 to simulate crowd pedestrian dynamics~\cite{helbing1995social}. After that, researchers made several improvements to the social force model. For example, to consider an extremely crowded situation where pedestrians might squeeze each other, the collision force for panic situation handling~\cite{helbing2000simulating} was added. To better describe the pedestrian-to-pedestrian interaction, new designs of collision avoidance~\cite{johansson2007specification, karamouzas2009predictive, zanlungo2011social} were proposed. Recently, to consider more complex scenarios that also include vehicle-to-pedestrian interaction and environmental context, the ordinary social force model was extended to become the multi-layer social force model~\cite{anvari2015modelling, pascucci2015modeling, rinke2017multi, johora2018modeling}. In the extension, while the original social force model is applied as the bottom layer to describe the explicit pedestrian motion, additional layers were added to represent higher-level operations such as decision-making and path planning in the complex scenarios.

\subsubsection{Vehicle Influence on Pedestrian}

How the pedestrian motion is influenced by vehicles is a specific aspect of social force based pedestrian motion modeling. 
A straightforward solution is to treat the vehicle as a static obstacle~\cite{zeng2017specification} or as an agent that is similar to the pedestrian, but with a different shape such as an elliptic-like shape~\cite{anvari2015modelling, yang2017agent, cheng2018modeling}. 
The vehicle influence can also be described as a combination of longitudinal and lateral effect~\cite{yang2018social} or by an influential point on an adjusted vehicle contour that considers some buffer space~\cite{yang2020social}. 
Another way of modeling vehicle influence is to modify the pedestrian's high-level operations in multi-layer social force models. For example, the pedestrian's decisions such as accelerating or decelerating can be determined based on the predicted conflict point between the pedestrian and the vehicle~\cite{pascucci2015modeling, pascucci2017discrete}. If the vehicle's reaction to the pedestrian is also considered, the game theory can be used to determine the priority~\cite{johora2018modeling}. 
There are some vehicle influence modeling approaches that are not specifically designed for social force models. Examples are the corporation-based trajectory planning~\cite{kabtoul2020towards} and the application of a binary function to determine whether or not the pedestrian is yielding to the vehicle~\cite{anderson2020off}. However, in general, the majority of the vehicle influence designs are proposed for social force models.

\subsubsection{Calibration and Evaluation}

In the early stage, calibrating social force models were relying on empirically tuning the model parameters and observing the simulation performance~\cite{helbing1995social, helbing2000simulating}. Later, the mainstream method has become using genetic algorithm~\cite{seer2014validating, dias2018calibrating, zeng2017specification} to minimize the difference between the simulated trajectories and the ground truth trajectories from certain datasets. Commonly used pedestrian-only datasets are ETH~\cite{pellegrini2009you} and UCY~\cite{lerner2007crowds}. Datasets including vehicles are SDD~\cite{robicquet2016learning}, HBS\cite{pascucci2017discrete}, CITR~\cite{yang2019top}, and DUT~\cite{yang2019top}. Since the social force model by design is very complex, using an evolutionary algorithm like genetic algorithm is still the most popular approach. 

Evaluating a social force model can be done either qualitatively or quantitatively.
For the models that only consider pedestrian-to-pedestrian interaction, the conventional way of evaluation is to check the fundamental diagram, i.e., the relationship between average pedestrian speed and pedestrian density, and to observe the self-organization phenomena such as lane formation~\cite{dong2019state}. When vehicle influence is considered, the evaluation can qualitatively check if the pedestrian behavior is reasonable in the simulation of vehicle-pedestrian interaction~\cite{anvari2015modelling, pascucci2015modeling, pascucci2017discrete, rinke2017multi, yang2018social}. If pedestrian trajectory data is available, metrics that quantitatively compute the difference between the simulated pedestrian motion and the motion in the datasets~\cite{johora2018modeling, yang2020social, chengtrajectory} can also be used. The most common metrics are the average displacement error (ADE) and the final displacement error (FDE).

\subsubsection{Application}

The primary application of the social force model is pedestrian motion simulation. For example, pedestrian dynamics in an evacuation scenario can be analyzed~\cite{helbing2000simulating}. Sometimes, it is also used for pedestrian tracking~\cite{luber2010people} and abnormal detection~\cite{mehran2009abnormal}. If the vehicle influence is considered, the model can be used to analyze the design of shared spaces~\cite{anvari2015modelling}. A social force model can also be combined with other methods for the task of pedestrian path prediction~\cite{zhang2020pedestrian}. In general, social force model provides basic descriptions for pedestrian motion, so it has a wide range of applications. 



\section{Problem Formulation}

\label{se:formulation}
The problem of modeling collective pedestrian motion under vehicle influence is transformed as finding a mathematical model that can be applied to each pedestrian and update the pedestrian's state as time evolves. The model associated with each pedestrian considers the state change of all surrounding agents (including vehicle) so that their influences are counted. The general form of such a pedestrian model $f_\text{ped}(.)$ is defined as:
\begin{equation}
    \mathbf{s}_{t+\Delta t}=f_\text{ped}(\mathbf{s}_t,v_\text{d},\mathbf{p}_\text{des},\mathbf{S}_t,\mathbf{E}_t),
    \label{eq:agent}
\end{equation}
where $\mathbf{s}_t=[p_\text{x}, p_\text{y}, v_\text{x}, v_\text{y}]^\text{T}$ is the state vector consisting of the position $p_\text{x}, p_\text{y}$ and the velocity $v_\text{x}, v_\text{y}$ in 2D Euclidean coordinates. The expression $f_\text{ped}(.)$ updates the pedestrian state $\mathbf{s}_t$ from the current time step $t$ to the next time step $t+\Delta t$, with $\Delta t$ being the time step interval. $v_\text{d}\in\mathbb{R}^+$ is the desired speed and $\mathbf{p}_\text{des}=[p_{x,\text{des}},p_{y,\text{des}}]^T$ is the destination position, which indicates that the pedestrian would like to reach $\mathbf{p}_\text{des}$ with speed $v_\text{d}$ in an ideal situation where there is no influence. $\mathbf{S}_t$ represents the states of all surrounding agents, including both pedestrians and vehicles. $\mathbf{E}_t$ is the environment context representing space layout and/or the locations of static obstacles. 

The collective pedestrian motion under vehicle influence in a specific scenario is defined as a process:
\begin{equation}
    \mathbb{P}:\mathbb{C}\rightarrow\{\boldsymbol{\tau}^i\},
    \label{eq:process}
\end{equation}
where $\mathbb{C}$ is the scenario configuration that specifies how the pedestrians are influenced by the vehicle. $\boldsymbol{\tau}^i$ is the generated trajectory of pedestrian $i$, which is a sequence of states $\mathbf{s}_t^i$ starting from the starting time $t=0$ to the end time $t=t_\text{end}$ with time step interval $\Delta t$. The process $\mathbb{P}$ outputs a set of all pedestrian trajectories $\{\boldsymbol{\tau}^i\}$, which represents the collective pedestrian motion. Algorithm~\ref{al:simulation} details the process $\mathbb{P}$. 
Generally speaking, the motion of each pedestrian $i$ is generated by applying the pedestrian model $f_\text{ped}^i(.)$, which is initialized with $\mathbf{s}_0^i$, $v_\text{d}^i$, and $\mathbf{p}_\text{des}^i$, under the influence of the vehicle. The vehicle motion is generated using a prescribed policy $P_\text{veh}$. $P_\text{veh}$ is a high-level representation that includes everything necessary to generate the vehicle motion. For example, the vehicle motion can be exactly the same as the motion in a data sample or generated by applying a path following controller to follow a pre-defined reference path.

\begin{algorithm}
\label{al:simulation}
\SetAlgoLined
\KwInput{Scenario $\mathbb{C}=(\{\mathbf{s}_0^i\},\{v_\text{d}^i\},\{\mathbf{p}_\text{des}^i\},P_\text{veh})$}
\KwOutput{Collective pedestrian trajectories $\{\boldsymbol{\tau}^i\}$.}

    \For{each pedestrian $i$}{
        initialize $f_\text{ped}^i(.)$ with $\mathbf{s}_0^i$, $v_\text{d}^i$, and $\mathbf{p}_\text{des}^i$\;
    }
    initialize the vehicle using $P_\text{veh}$\;
    $t=0$\;
    \While{$t<t_\text{end}$}{
        \For{each pedestrian $i$}{
            $\mathbf{s}_{t+\Delta t}=f_\text{ped}(\mathbf{s}_t,v_\text{d},\mathbf{p}_\text{des},\mathbf{S}_t,\mathbf{E}_t)$; 
            \tcp{equation~\ref{eq:agent}} 
        }
        use $P_\text{veh}$ to update the vehicle motion\;
        $t=t+\Delta t$\;
    }
    \caption{Process $\mathbb{P}$}
\end{algorithm}

In this work, the problem is defined as finding a pedestrian motion model $f_\text{ped}(.)$ such that $\forall \mathbb{C} \in \mathcal{F}_\mathbb{C}$, the collective pedestrian motion $\{\boldsymbol{\tau}^i\}$ generated by the process $\mathbb{P}$ is effective. 
$\mathcal{F}_\mathbb{C}$ refers to a set of fundamental scenarios representing a variety of collective pedestrian motion under vehicle influence (details in section~\ref{subse:eval_sim} and illustrated in figure~\ref{fig:fund_scenarios}).
The word \textit{effective} has two interpretations. First, the pedestrian motion must be collision-free. Second, the pedestrian motion patterns should be as realistic as possible.

\section{Modeling Process}
\label{se:design_flow}

\begin{figure}
    \centering
    \includegraphics[width=0.7\linewidth]{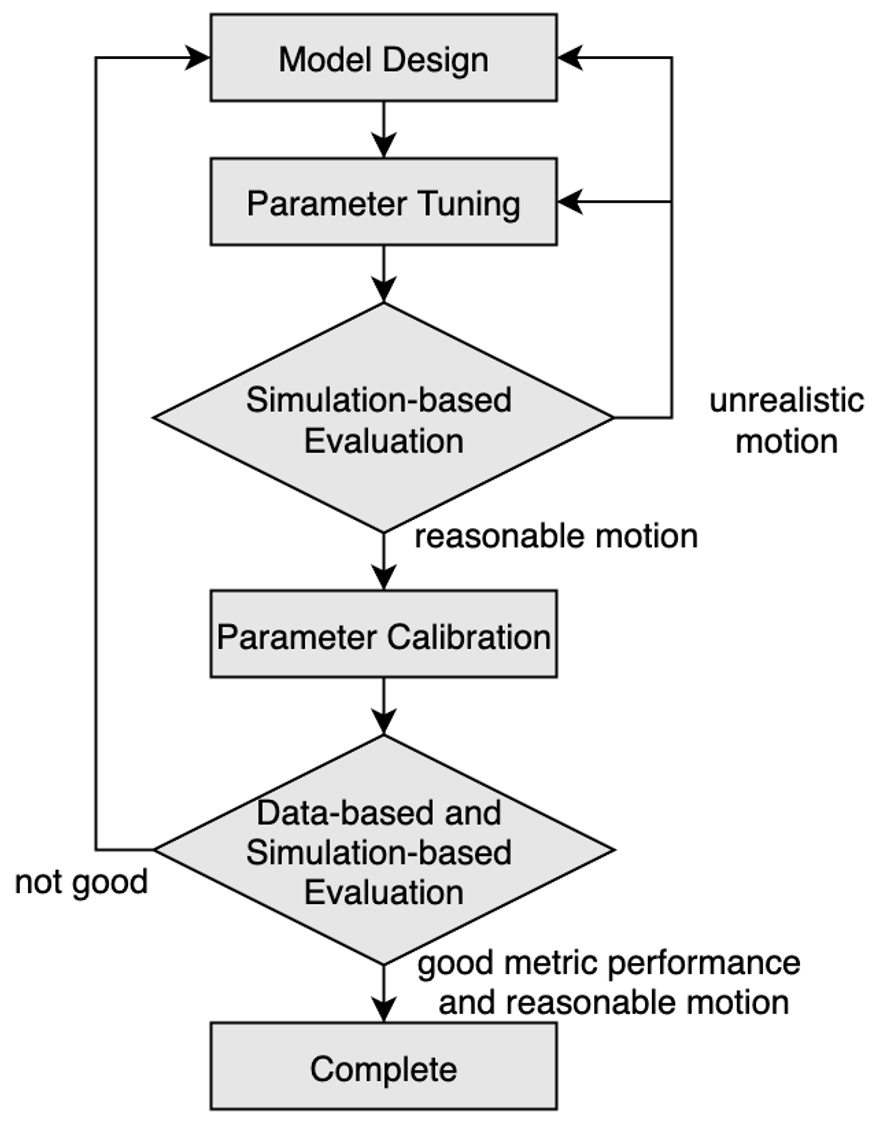}
    \caption{Process of designing and evaluating a pedestrian motion model.}
    \label{fig:design_flow}
\end{figure}

A modeling process was proposed to to systematically design and evaluate the pedestrian motion model $f_\text{ped}(.|\boldsymbol{\theta})$, where $\boldsymbol{\theta}$ is the model parameters. This is an iterative process including model design, parameter adjustment, and model evaluation, as shown in figure~\ref{fig:design_flow}. It takes the advantages of both the data-based evaluation and the simulation-based evaluation. The major steps in the process are explained as follows:

\begin{itemize}
    
    \item \textit{Prototype Design}: This is the first step. It refers to analytically developing the pedestrian motion model $f_\text{ped}(.|\boldsymbol{\theta})$ to meet the requirement. The overall design needs to consider the core characteristics of the pedestrian motion, while the parameters $\boldsymbol{\theta}$ are responsible for the details of the motion. Adjusting the values of $\boldsymbol{\theta}$ shouldn't dramatically change the model performance.
    
    \item \textit{Parameter Tuning}: Once the prototype is ready, the next step is to manually tune the parameters $\boldsymbol{\theta}$. The parameter tuning is conducted with the help of observing the pedestrian behavior generated by the process $\mathbb{P}$ with different $\mathbb{C}$ (equation~\ref{eq:process}). The value of each parameter must be justifiable according to the model design. For example, a parameter for a maximum force magnitude must be positive.
    
    \item \textit{Simulation-based Evaluation}: The next step is to evaluate the tuned model over all fundamental scenarios $\mathbb{C}\in\mathcal{F}_\mathbb{C}$ using process $\mathbb{P}$. If all the pedestrian motion is realistic enough, then the modeling process enters the parameter calibration. Otherwise, the unrealistic pedestrian motion needs to be analyzed and the possible reasons need to be identified. Based on the reasons, the modeling process either goes back to the parameter tuning and adjust the parameters again or goes further back to the prototype design and modify the model. The details of simulation-based evaluation can be found in section~\ref{subse:eval_sim}.
    
    \item \textit{Parameter Calibration}: This step further refines the model by calibrating the parameters $\boldsymbol{\theta}$ utilizing the pedestrian datasets. The datasets covers several scenarios of vehicle-pedestrian interaction (see section~\ref{subse:dataset}). If a fitness function $J(\boldsymbol{\theta})$ is defined to describe the similarity between the motion generated by the model and the corresponding ground truth motion, then the calibration is to find $\boldsymbol{\theta}^*=\arg\min_{\boldsymbol{\theta}}J(\boldsymbol{\theta})$. The details of the parameter calibration is presented in section~\ref{se:calibration}.
    
    \item \textit{Data-based and Simulation-based Evaluation}: After calibration, the model needs to be evaluated again. 
    In this step, besides the simulation-based evaluation, data-based evaluation is also conducted as a supplement. The data-based evaluation utilizes a couple of metrics to quantitatively assess the model performance over data (details in section~\ref{subse:eval_data}). If both evaluations show good results, it means the model performance is fairly good and the modeling process is completed. If not, possible reasons should be identified by analyzing the evaluation results and based on which the model prototype needs to be improved again. 
    
\end{itemize}


The proposed SG-SFM was successfully developed by exactly following the above process. This process is also applicable to any other models that deal with collective pedestrian motion in a similar situation.

\section{Sub-Goal Social Force Model}
\label{se:pedestrian_model}

This section describes the details of the sub-goal social force model (SG-SFM). The description starts with motion dynamics, then goes to the two categories of sub-forces, repulsive force and navigational force, and finally physical constraints of pedestrian motion. 


\subsection{Motion Dynamics}
The motion of a pedestrian follows a point-mass dynamics. Let's denote the pedestrian's position as $\mathbf{p}=[p_\text{x}, p_\text{y}]^\text{T} \in \mathbb{R}^2$, velocity as $\mathbf{v}=[v_\text{x}, v_\text{y}]^\text{T}\in \mathbb{R}^2$, and acceleration as $\mathbf{a}=[a_\text{x}, a_\text{y}]^\text{T}\in \mathbb{R}^2$, then the dynamics model is expressed as:
\begin{equation}
    \frac{d^2\mathbf{p}}{dt^2}=\frac{d\mathbf{v}}{dt}=\mathbf{a}.
	\label{eq:dynamics}
\end{equation}
The above equation is solved by the time-stepping of Smart-Euler with time step $\Delta t$. The acceleration $\mathbf{a}$ is obtained by:
\begin{equation}
    \mathbf{a}=\frac{\mathbf{F}_\text{total}}{m},
\end{equation}
where $m$ is the mass of the pedestrian. $\mathbf{F}_\text{total}\in \mathbb{R}^2$ is the total force that acts upon the point-mass dynamics, which represents the pedestrian's action to accomplish certain tasks like reaching the destination or reacting to the change of environment.

The proposed approach does not follow the traditional design of splitting the total force $\mathbf{F}_\text{total}$ into several categories such as destination force, pedestrian force, and vehicle force. Instead, the total force $\mathbf{F}_\text{total}$ is spitted into only two categories, repulsive force $\mathbf{F}_\text{rep}$ and navigational force $\mathbf{F}_\text{nav}$:
\begin{equation}
    \mathbf{F}_\text{total}=\mathbf{F}_\text{rep}+\mathbf{F}_\text{nav}.
\end{equation}
This is under the belief that the pedestrian's action should be either \textit{reactive} (repulsive force) or \textit{proactive} (navigational force), or a combination of both. 
Therefore, these two forces can be expressed as $\mathbf{F}_\text{rep}=f_\text{rep}(\mathbf{s},\mathbf{S},\mathbf{E})$ and $\mathbf{F}_\text{nav}=f_\text{nav}(\mathbf{s},\mathbf{S},\mathbf{E}|v_\text{d},\mathbf{p}_\text{des})$, 
where $\mathbf{s}$ is the state of ego pedestrian and $\mathbf{S}$ is a list of states of surrounding pedestrians and vehicles. For simplicity, the time notation $t$ is dropped. Repulsive force $\mathbf{F}_\text{rep}$ is reactive, so it does not depend on the desired speed $v_\text{d}$ and the destination $\mathbf{p}_\text{des}$. Navigational force $\mathbf{F}_\text{nav}$ is proactive, so it depends on $v_\text{d}$ and $\mathbf{p}_\text{des}$. The environment context $\mathbf{E}$ mainly includes information of the static obstacles. The space layout is omitted here by assuming a large open shared space, as it can be easily added if necessary.

\subsection{Repulsive Force}
Repulsive force $\mathbf{F}_\text{rep}$ keeps the ego pedestrian certain distance away from any vehicle, pedestrian, or obstacle. Therefore, it is a summation of all individual repulsive forces:
\begin{equation}
    \mathbf{F}_\text{rep}=\sum_i\mathbf{F}_\text{rep}^{\text{veh},i}+\sum_i\mathbf{F}_\text{rep}^{\text{ped},i}+\sum_i\mathbf{F}_\text{rep}^{\text{obs},i},
\end{equation}
where $\mathbf{F}_\text{rep}^{\text{veh},i}$ represents a repulsive force from a vehicle, and similarly, $\mathbf{F}_\text{rep}^{\text{ped},i}$ from a pedestrian and $\mathbf{F}_\text{rep}^{\text{obs},i}$ from an obstacle. Each force consists of a force direction (a unit vector) and a force magnitude.
For clarity, the index $i$ is dropped in the following description.

\begin{figure}
    \centering
    \includegraphics[width=\linewidth]{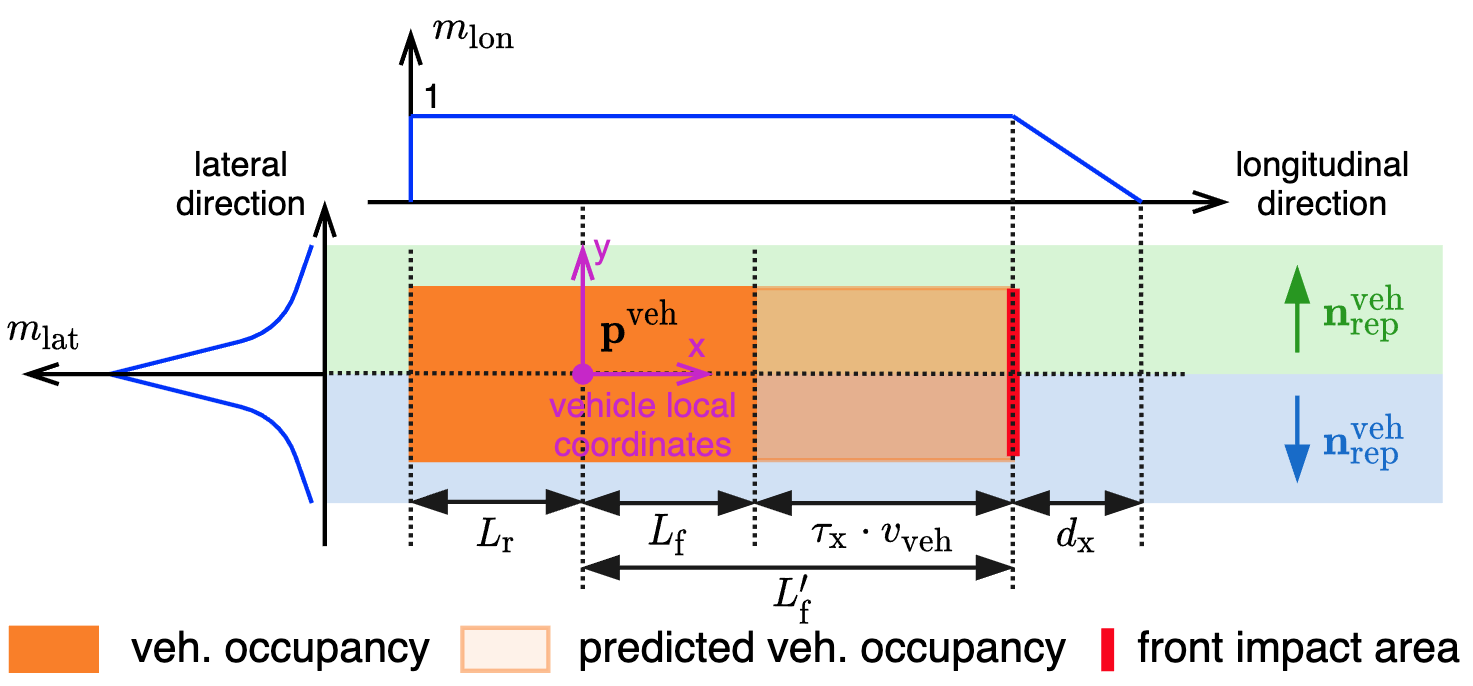}
    \caption{Illustration of the repulsive force from a vehicle. The force direction ($\mathbf{n}_\text{rep}^\text{veh}$) points to either positive or negative y axis (green and blue arrow). The force magnitude is the multiplication of the longitudinal magnitude ($m_\text{lon}$) and the lateral magnitude ($m_\text{lat}$). See detailed description in the text.}
    \label{fig:rep_veh}
\end{figure}

\subsubsection{Vehicle Repulsive Force}
Since the vehicle is the most safety critical agent that has influence on pedestrians, we have a specific design of its repulsive force, which has the form: 
\begin{equation}
    \mathbf{F}_\text{rep}^{\text{veh}}=m_\text{rep}^\text{veh}\cdot \mathbf{n}_\text{rep}^\text{veh},
\end{equation}
where $\mathbf{n}_\text{rep}^\text{veh}$ is the force direction and $m_\text{rep}^\text{veh}$ is the force magnitude. Figure~\ref{fig:rep_veh} illustrates how they are designed. 

The force direction $\mathbf{n}_\text{rep}^\text{veh}$ has two choices. In the vehicle's local coordinates, assuming the vehicle's center position $\mathbf{p}^\text{veh}$ is at origin and the vehicle is heading to positive x axis, the force direction $\mathbf{n}_\text{rep}^\text{veh}$ points to either positive y axis (ego pedestrian in the upper half plane, as shown in the green area of figure~\ref{fig:rep_veh}) or negative y axis (ego pedestrian in the lower half plane, as shown in the blue area of figure~\ref{fig:rep_veh}). 

The force magnitude $m_\text{rep}^\text{veh}$ is the multiplication of lateral magnitude $m_\text{lat}$ and longitudinal magnitude $m_\text{lon}$:
\begin{equation}
    m_\text{rep}^\text{veh}=m_\text{lat}\cdot m_\text{lon},
\end{equation}
which is also illustrated in figure~\ref{fig:rep_veh}.
The lateral magnitude $m_\text{lat}$ decreases as the lateral distance $d_\text{lat}$ between the vehicle and the ego pedestrian increases:
\begin{equation}
    \label{eq:mag_veh}
    m_\text{lat}=M_\text{veh}\cdot \exp(-\beta_\text{veh}\cdot d_\text{lat}),
\end{equation}
where $M_\text{veh},\beta_\text{veh}$ are parameters for the exponential function. 
The longitudinal magnitude $m_\text{lon}$ is calculated as follow:
\begin{equation}
    m_\text{lon}=
    \begin{cases}
        1, & \text{if } -L_\text{r} < p_\text{x}^\text{ego} < L_\text{f}^\prime \\
        1-\frac{p_\text{x}^\text{ego}-L_\text{f}^\prime}{d_\text{x}}, & \text{if } L_\text{f}^\prime < p_\text{x}^\text{ego} < L_\text{f}^\prime+d_\text{x} \\
        0, & \text{otherwise}
    \end{cases}
\end{equation}
where $p_\text{x}^\text{ego}$ is the ego pedestrian x position in vehicle's coordinates. $L_\text{r}$ is the vehicle's center-to-rear distance. $L_\text{f}^\prime=L_\text{f}+\tau_\text{x}\cdot v_\text{veh}$ is the distance from the vehicle center to front impart area (the red solid line in figure~\ref{fig:rep_veh}). $L_\text{f}^\prime$ considers both the center-to-front distance $L_\text{f}$ and the predicted motion of the vehicle, which is the longitudinal vehicle speed $v_\text{veh}$ multiplied by a parameter $\tau_\text{x}$. Beyond front impact area, a buffer distance $d_\text{x}$ is added to gradually reduce $m_\text{lon}$ to from $1$ to $0$.

This design mainly considers the vehicle's future motion and the pedestrian's unwillingness of staying inside the vehicle's driving path. 

\subsubsection{Pedestrian Repulsive Force}
\begin{figure}
    \centering
    \includegraphics[width=0.8\linewidth]{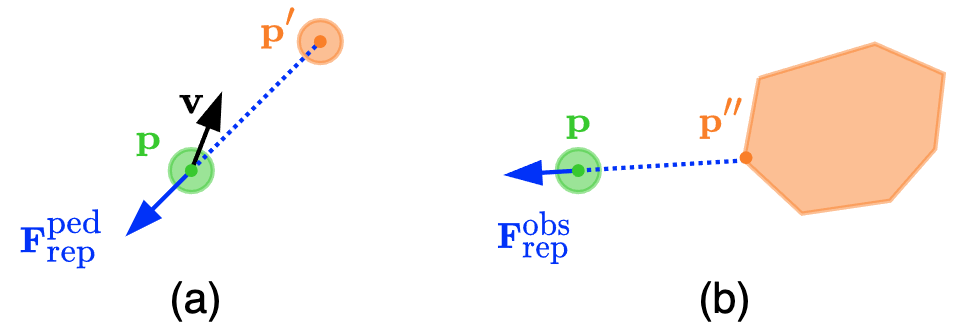}
    \caption{Repulsive forces from a surrounding pedestrian ($\mathbf{F}_\text{rep}^\text{ped}$) and an obstacle ($\mathbf{F}_\text{rep}^\text{obs}$). (a) $\mathbf{p}^\prime$ is the position of the surrounding pedestrian. (b) $\mathbf{p}^{\prime\prime}$ is the influential point of the obstacle.}
    \label{fig:rep_ped_obs}
\end{figure}

Pedestrian repulsive force considers the geometric relationship between the ego pedestrian and a surrounding pedestrian, as illustrated in figure~\ref{fig:rep_ped_obs}(a). It has the form:
\begin{equation}
    \mathbf{F}_\text{rep}^{\text{ped}}=m_\text{rep}^\text{ped}\cdot A_\text{rep}^\text{ped}\cdot \mathbf{n}_\text{rep}^\text{ped}.
\end{equation}
In addition to force direction $\mathbf{n}_\text{rep}^\text{ped}$ and force magnitude $m_\text{rep}^\text{ped}$, the anisotropy $A_\text{rep}^\text{ped}$ of pedestrian repulsion is also considered, which describes the change of repulsion when the interacting pedestrian comes from different angle.

The force direction $\mathbf{n}_\text{rep}^\text{ped}$ is a unit vector pointing from the surrounding pedestrian's position $\mathbf{p}^\prime$ to the ego pedestrian's position $\mathbf{p}$. 
The force magnitude $m_\text{rep}^\text{ped}$ decreases as the distance between two pedestrians increases:
\begin{equation}
    \label{eq:mag_ped}
    m_\text{rep}^\text{ped}=M_\text{ped}\cdot \exp(-\beta_\text{ped}\cdot (||\mathbf{p}
^\prime-\mathbf{p}||-2 R_\text{ped})), 
\end{equation}
where $M_\text{ped},\beta_\text{ped}$ are parameters.
$R_\text{ped}$ is the pedestrian's radius which allows two pedestrians to slightly push and squeeze each other. 
The anisotropy $A_\text{rep}^\text{ped}$ is calculated as follow:
\begin{equation}
    A_\text{rep}^\text{ped}=\alpha_\text{ped}+(1-\alpha_\text{ped})\cdot\frac{1+\cos||\langle\mathbf{v},\mathbf{p}^\prime-\mathbf{p}\rangle||}{2},
\end{equation}
 where $\mathbf{v}$ is the ego pedestrian's velocity and $\alpha_\text{ped}$ is a parameter. The anisotropy allows a pedestrian from behind has less effect than a pedestrian in front.

\subsubsection{Obstacle Repulsive Force}
Obstacles usually have irregular shapes. To describe the effect from an obstacle, we select the closest point $\mathbf{p}^{\prime\prime}$ on the obstacle as the influential point, as shown in figure~\ref{fig:rep_ped_obs}(b), then the obstacle repulsive force can be described in a similar way as the pedestrian repulsive force:
\begin{equation}
    \mathbf{F}_\text{rep}^\text{obs}=    m_\text{rep}^\text{obs}\cdot\mathbf{n}_\text{rep}^\text{obs},
\end{equation}
where $\mathbf{n}_\text{rep}^\text{obs}$ is the force direction pointing from the influential point $\mathbf{p}^{\prime\prime}$ to the ego pedestrian's position $\mathbf{p}$ and the magnitude:
\begin{equation}
    m_\text{rep}^\text{obs}=M_\text{obs}\cdot \exp(-\beta_\text{obs}\cdot(||\mathbf{p}^{\prime\prime}-\mathbf{p}||-R_\text{ped}))
\end{equation}

\begin{figure}
    \centering
    \includegraphics[width=0.8\linewidth]{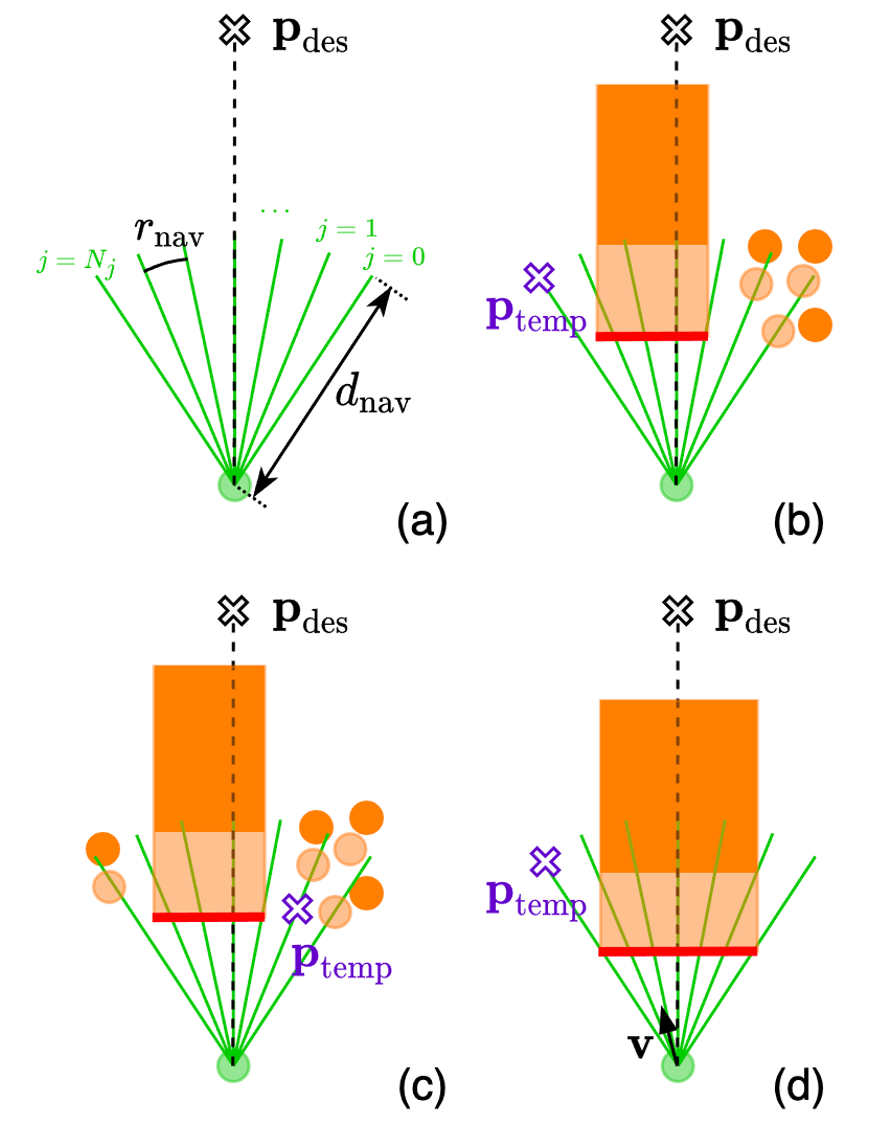}
    \caption{The process of temporary destination selection. (a) The creation of candidate directions (green solid line) centered on the direction to the final destination (black dashed line). (b)(c)(d) The evaluation of each direction and the determination of temporary destination $\mathbf{p}_\text{temp}$ (see algorithm~\ref{al:temp_goal} for detail). The rectangle represents the vehicle occupancy. The circle represents the the surrounding pedestrian occupancy. Dark orange color indicates current occupancy while light orange color indicates the predicted occupancy. The red bold solid line is the vehicle front impact area (the same as defined in figure~\ref{fig:rep_veh}).  }
    \label{fig:temp_goal}
\end{figure}

\subsection{Navigational Force}
Navigational force describes the pedestrian's navigation around the surrounding agents while moving toward the final destination $\mathbf{p}_\text{des}$. It drives the ego pedestrian to change the walking velocity $\mathbf{v}$ to a target walking velocity $\mathbf{v}_\text{tar}$:
\begin{equation}
    \mathbf{F}_\text{nav}=K_\text{nav}\cdot(\mathbf{v}_\text{tar}-\mathbf{v}),
\end{equation}
where $K_\text{nav}$ is a parameter. The target velocity $\mathbf{v}_\text{tar}$ also changes at every time step, because the interaction situation changes as time evolves. This is achieved by identifying a temporary destination $\mathbf{p}_\text{temp}$ for the pedestrian to reach. The pedestrian also has a pre-defined desired walking speed $v_\text{d}$, which is the preferred speed when there is no interaction. With both $\mathbf{p}_\text{temp}$ and $v_\text{d}$, the target velocity $\mathbf{v}_\text{tar}$ can be obtained by: 
\begin{equation}
	\mathbf{v}_\text{tar}=v_\text{d}\cdot\frac{\mathbf{p}_\text{temp}-\mathbf{p}}{\sqrt{||\mathbf{p}_\text{temp}-\mathbf{p}||^2+\sigma^2}},
\end{equation}
where $\sigma$ is a parameter that reduces the magnitude of $\mathbf{v}_\text{tar}$ when the pedestrian's position $\mathbf{p}$ is getting close to $\mathbf{p}_\text{temp}$~\cite{yang2020social}.


\begin{algorithm}[t]
\caption{Determining Temporary Destination}
\label{al:temp_goal}
    \SetAlgoLined
    \KwInput{ego pedestrian state $\mathbf{s}$, states of surrounding agents $\mathbf{S}$, destination position $\mathbf{p}_\text{des}=(\phi_\text{des},d_\text{des})$}
    \KwOutput{temporary destination position $\mathbf{p}_\text{temp}=(\phi_{j^*},d_{j^*})$}
    \KwParameters{$N_j$, $r_\text{nav}$, $d_\text{nav}$, $R_\text{ped}$} 
    
    set of candidate directions $\mathbb{S}=\{j|0\le j\le N_j, j\in \mathbb{N}\}$\;
    \For{$j\in \mathbb{S}$}{
        $\phi_j=\phi_\text{des}+(j-\frac{N_j}{2})\cdot r_\text{nav}$ \tcp*{see figure~\ref{fig:temp_goal}(a)}
        \uIf{no obstruction in direction $j$}{
            $d_j=d_\text{nav}$\;
            $c_j=\text{None}$\;
        }
        \Else{
            $d_j=||\mathbf{p}-\mathbf{p}_j||-R_\text{ped}$\;
            \uIf{obstructed by vehicle front impact area}{
                $c_j=\text{Front}$\;
            }
            \Else{
                $c_j=\text{Other}$\;
            }
        }
    }
    
    set of passable directions $\mathbb{S}^\text{p}=\{j|d_j=d_\text{nav},\forall j \in \mathbb{S}\}$\;
    \uIf{$\mathbb{S}^\text{p}\ne\emptyset$}{
        \tcp{see figure~\ref{fig:temp_goal}(b)}
        $j^*=\arg\min_j |\phi_j-\phi_\text{des}|, \forall j\in\mathbb{S}^\text{p} $\;
    }
    \Else{
        set of directions not facing vehicle front $\mathbb{S}^\text{n}=\{j|c_j\ne\text{Front},\forall j \in \mathbb{S}\}$\;
        \uIf{$\mathbb{S}^\text{n}\ne\emptyset$}{
            \tcp{see figure~\ref{fig:temp_goal}(c)}
            $j^*=\arg\min_j |\phi_j-\phi_\text{des}|, \forall j\in\mathbb{S}^\text{n} $\;
        }
        \Else{
            \tcp{see figure~\ref{fig:temp_goal}(d)}
            get ego pedestrian's velocity angle $\phi_\text{ego}$\;
            \uIf{$|\phi_\text{ego}-\phi_0|<|\phi_\text{ego}-\phi_{N_j}|$}{
                $j^*=0$\;
            }
            \Else{
                $j^*=N_j$\;
            }
        }
    }

\end{algorithm}

To determine the temporary destination $\mathbf{p}_\text{temp}$, the ego pedestrian considers both the current and the expected positions of surrounding agents with respect to the direction from the ego pedestrian to the final destination $\mathbf{p}_\text{des}$. This process is illustrated in figure~\ref{fig:temp_goal} and presented in algorithm~\ref{al:temp_goal}. Generally speaking, the temporary destination $\mathbf{p}_\text{temp}$ is determined with respect to the ego pedestrian's local polar coordinates. The final destination $\mathbf{p}_\text{des}$, along with the states of surrounding agents $\mathbf{S}$ as the input to the algorithm, is rewritten as $\mathbf{p}_\text{des}=(\phi_\text{des},d_\text{des})$, where $\phi_\text{des}$ is the direction and $d_\text{des}$ is the distance.
In algorithm~\ref{al:temp_goal}, the first step is to create a set of $N_j$ candidate directions $\mathbb{S}$ centered around $\phi_\text{des}$. Then for each direction $\phi_j$, the navigation range $d_j$ and the obstruction type $c_j$ are calculated. Finally, the temporary destination $\mathbf{p}_\text{temp}$ is found based on $\phi_j$, $d_j$, and $c_j$: 
\begin{equation}
    \mathbf{p}_\text{temp}=(\phi_{j^*},d_{j^*}).
\end{equation}
There are some parameters in algorithm~\ref{al:temp_goal}. $N_j$ is the number of candidate directions, $r_\text{nav}$ is angle between two adjacent directions, $d_\text{nav}$ is the maximum navigation range, and $R_\text{ped}$ is the radius of pedestrian.

\subsection{Total Force Limit}

\begin{figure}
    \centering
    \includegraphics[width=0.8\linewidth]{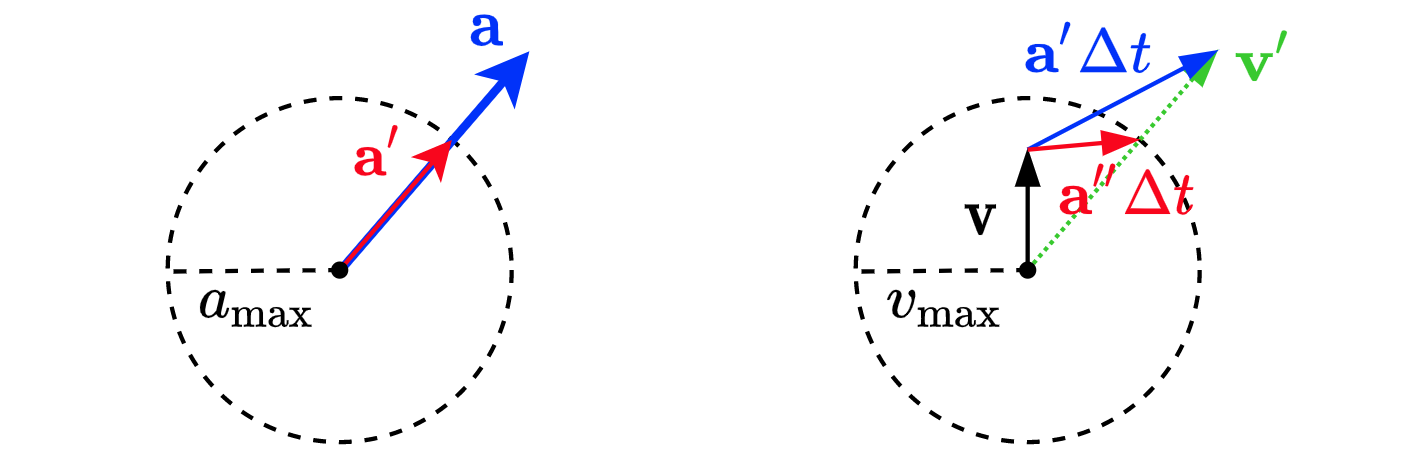}
    \caption{Illustration of how the total force $\mathbf{F}_\text{total}$ is limited by modifying the corresponding acceleration. $\mathbf{a}^{\prime}$: modified acceleration based on the acceleration limit $a_\text{max}$. $\mathbf{a}^{\prime\prime}$: modified acceleration based on the velocity limit $v_\text{max}$.}
    \label{fig:constraint}
\end{figure}

The total force $\mathbf{F}_\text{total}$ is limited according to the maximum acceleration $a_\text{max}$ and maximum speed $v_\text{max}$ of the pedestrian, as the pedestrian motion cannot be changed abruptly in reality.
As shown in figure~\ref{fig:constraint}, $\mathbf{F}_\text{total}$ is firstly interpreted as an acceleration vector $\mathbf{a}=\mathbf{F}_\text{total}/m$. If $||\mathbf{a}||>a_\text{max}$, the acceleration vector is modified as:
\begin{equation}
    \mathbf{a}^\prime=a_\text{max}\cdot \frac{\mathbf{a}}{||\mathbf{a}||}.
\end{equation}
Then the expected new velocity vector $\mathbf{v}^\prime=\mathbf{v}+\mathbf{a}^\prime\cdot\Delta t$ is calculated, where $\mathbf{v}$ is the current velocity vector. If $||\mathbf{v}^\prime||>v_\text{max}$, we further modify the acceleration vector: 
\begin{equation}
    \mathbf{a}^{\prime\prime}=\frac{1}{\Delta t}\cdot(v_\text{max}\cdot \frac{\mathbf{v}^\prime}{||\mathbf{v}^\prime||}-\mathbf{v}).
\end{equation}
Finally the total force is updated as:
\begin{equation}
    \mathbf{F}_\text{total}=m\cdot\mathbf{a}^{\prime\prime}
\end{equation}

\section{Parameter Tuning and Calibration}
\label{se:calibration}


The model parameters are adjusted by both tuning and calibration to improve the model performance. Table~\ref{tab:parameters} shows the parameters to be tuned and calibrated, respectively. \textit{Parameter tuning} does not use dataset. All parameter values are manually adjusted based on the understanding of the prototype model and the feedback from the pedestrian behavior in simulation. The effectiveness of the manually tuned parameters is verified by the simulation-based evaluation. As a result, the model can achieve coarse but reasonable pedestrian motion. \textit{Parameter calibration} uses dataset. It further adjusts part of the parameters that are tuned in the previous step. These parameters are responsible for the pedestrian's detailed behavior. For example, how the magnitude of a repulsive force decreases as the distance increases (the exponential relationship in equation~\ref{eq:mag_veh} or equation~\ref{eq:mag_ped}). This work applies the genetic algorithm to conduct parameter calibration.

\begin{table}
    \centering
    \caption{Summary of parameters in the sub-goal social force model}
    \label{tab:parameters}
    \begin{tabular}{|l|l|l|}
        \hline
        \textbf{Category} & \textbf{Param. for Tuning} & \textbf{Param. for Cal.} \\ \hline
        Ped. Repulsive Force & $M_\text{ped}, \beta_\text{ped}, \alpha_\text{ped}$ & $\beta_\text{ped}$  \\ \hline
        Veh. Repulsive Force & $M_\text{veh}, \beta_\text{veh}, \tau_\text{x}, d_\text{x}$ & $\beta_\text{veh}, \tau_\text{x}, d_\text{x}$ \\ \hline
        Navigational Force & $K_\text{nav}, \sigma, N_j, r_\text{nav}, d_\text{nav}$ & $K_\text{nav}, N_j, d_\text{nav}$\\ \hline
    \end{tabular}
\end{table}

\subsection{Data Preparation}
\subsubsection{Creating Samples}
Each dataset is partitioned into a certain number of scenarios. A scenario is extracted to include continuous and complete interactions between pedestrians and vehicles. Each scenario is further partitioned into individual data samples. In each sample, a pedestrian is selected from the scenario as the ego pedestrian, while other pedestrians and the vehicles are treated as the surrounding agents. A scenario of $n$ pedestrians can be partitioned into $n$ individual samples. Note that the trajectory length of each individual sample is varied. This is different from the common setting for the dataset containing pedestrian-only scenarios~\cite{pellegrini2009you}. The varied length can ensure that the ego pedestrian goes through complete interaction with the vehicle.  

\subsubsection{Destination}
The SG-SFM requires a destination $\mathbf{p}_\text{des}$ as the input. The destination is estimated by extending the ground truth trajectory of the ego pedestrian:
\begin{equation}
    \mathbf{p}_\text{des}=\mathbf{p}_k^\text{gt}+l_\text{des}\cdot (\mathbf{p}_k^\text{gt}-\mathbf{p}_0^\text{gt}),
\end{equation}
where $\mathbf{p}_0^\text{gt}$ and $\mathbf{p}_k^\text{gt}$ are the first position and the last position of the trajectory, respectively. $l_\text{des}$ is the extended length. In this work, we choose $l_\text{des}=5m$. 

\subsubsection{Desired Speed}
Another required input to the SG-SFM is the desired walking speed $v_\text{d}$. The ego pedestrian in each data sample should have his/her own $v_\text{d}$. The $v_\text{d}$ is obtained in two steps. First, the portion of the ego trajectory at which the pedestrian is walking is identified. A walking status is determined based on if the pedestrian's speed is larger than a threshold $v_\text{walking}$. Then, we average all the walking speed values to obtain $v_\text{d}$. In this work we choose $v_\text{walking}=0.8m/s$.

\subsection{Genetic Algorithm}
\label{subse:ga}
The genetic algorithm is a class of evolutionary algorithms commonly used for the parameter calibration of social force based models. The algorithm imitates the process of natural selection. Basic operators in the genetic algorithm include mutation, crossover, and selection. It is a randomized search algorithm that explores a wide range of solutions while in the meantime applicable to a variety of optimization problems.

The parameters to be calibrated are combined into a vector $\boldsymbol{\theta}=[\beta_\text{ped},\beta_\text{veh}, \tau_\text{x}, d_\text{x},K_\text{nav}, N_j, d_\text{nav}]$ as the genetic representation of the solution for the genetic algorithm. The fitness function is defined as:
\begin{equation}
    f_\text{fit}(\boldsymbol{\theta})=\frac{1}{n}\sum_{j=1}^{n}f_\text{fit}^j(\boldsymbol{\theta}),
\end{equation}
where $n$ is the total number of data samples and $f_\text{fit}^j(\boldsymbol{\theta})$ is the fitness value of a particular data sample. For each sample $j$:
\begin{equation}
    f_\text{fit}^j(\boldsymbol{\theta})=\frac{1}{k}\sum_{i=1}^{k}||\mathbf{p}_i^\text{sim}-\mathbf{p}_i^\text{gt}||,
\end{equation}
which is the average displacement error between the simulated trajectory $\mathbf{p}^\text{sim}=[\mathbf{p}_0^\text{sim},\mathbf{p}_1^\text{sim}, \cdots, \mathbf{p}_k^\text{sim}]$ and the ground truth trajectory $\mathbf{p}^\text{gt}=[\mathbf{p}_0^\text{gt},\mathbf{p}_1^\text{gt}, \cdots, \mathbf{p}_k^\text{gt}]$. The simulated trajectory $\mathbf{p}^\text{sim}$ is obtained by simulating the ego pedestrian in a data sample using $f_\text{ped}(.|\boldsymbol{\theta})$, initialized with the first point in $\mathbf{p}^\text{gt}$, i.e., $\mathbf{p}_0^\text{sim}=\mathbf{p}_0^\text{gt}$. In the process of the simulation, the states of surrounding agents always use the ground truth values. 

\subsection{Calibration}
\label{subse:uni_vs_group}

Two types of calibration are conducted in this work. The first type is called \textit{universal calibration}. It searches for a unique best parameter vector $\boldsymbol{\theta}^*$ by assuming that all pedestrians are homogeneous. This assumption is acceptable because this work focuses on collective pedestrian motion instead of the individual motion of a particular pedestrian.

The second type is called \textit{group calibration}. In this case, the previous assumption is slightly modified to allow different motion characteristics~\cite{alahi2017learning}.  Pedestrians are classified into different groups. A unique best parameter vector $\boldsymbol{\theta}_i^*$ will be found for each group $i$.

\begin{figure}
    \centering
    \includegraphics[width=0.65\linewidth]{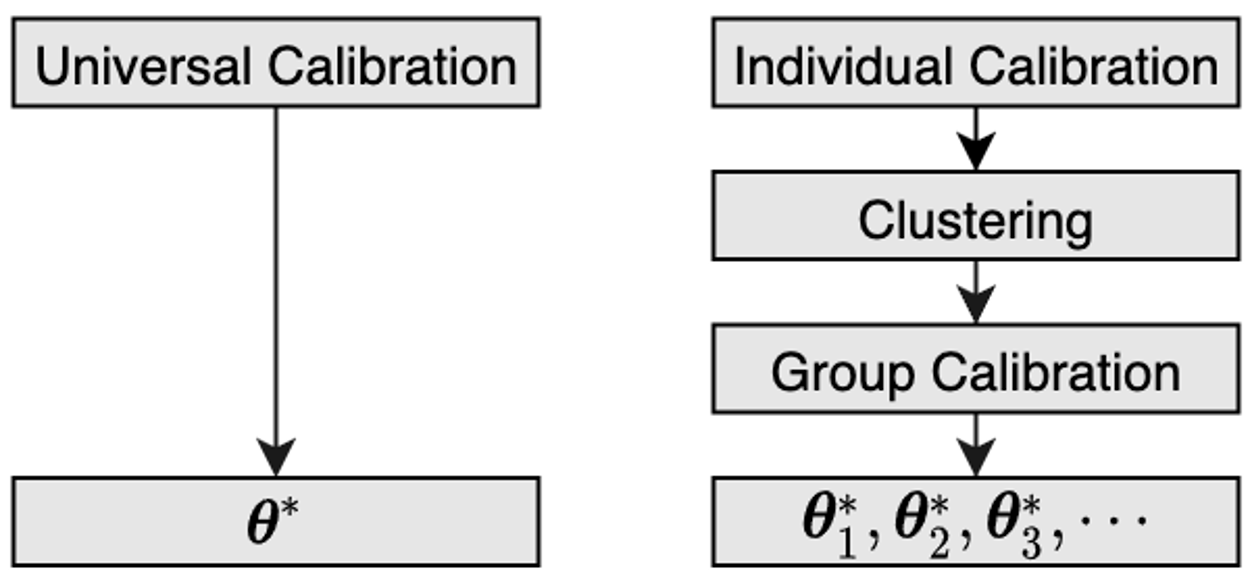}
    \caption{The process of universal calibration (left) and group calibration (right). Clustering applies K-means algorithm. See detail in section~\ref{subse:uni_vs_group}.}
    \label{fig:group_cal}
\end{figure}

Figure~\ref{fig:group_cal} compares the process of universal calibration and group calibration. The universal calibration directly applies the genetic algorithm. The output is exactly $\boldsymbol{\theta}^*$. The group calibration goes through a process of individual calibration, clustering, and group calibration. In the individual calibration, each pedestrian/sample is calibrated separately to obtain a preliminary parameter vector. The preliminary parameters of all the pedestrians are treated as the pedestrian motion features. Then the K-means algorithm~\cite{marutho2018determination} is used to cluster and analyze the motion features and consequently, the pedestrians are classified into different groups. Finally, the pedestrians in each group $i$ are calibrated together again by the genetic algorithm to obtain the best parameter vector $\boldsymbol{\theta}_i^*$.

\section{Evaluation}
\label{se:evaluation}


\subsection{Data-based Evaluation}
\label{subse:eval_data}
Data-based evaluation utilizes the scenarios in the dataset to quantitatively evaluates the model. It evaluates the quality of the simulated ego pedestrian trajectory $\mathbf{p}^\text{sim}$ in each data sample. $\mathbf{p}^\text{sim}$ is obtained in the same way as in the calibration (see section~\ref{subse:ga}). The following metrics are used for the evaluation:

\begin{itemize}

    \item \textit{Adjusted Average Displacement Error (aADE)}: 
    Average displacement error (ADE) is one of the most common metrics for pedestrian trajectory evaluation, especially for the trajectory obtained by prediction~\cite{sadeghian2019sophie}. In this work, the simulated trajectory are evaluated in a similar way, but with some modification. ADE is defined as the error between the simulated trajectory and the ground truth trajectory: $e^{\text{ADE}}=\frac{1}{k}\sum_{i=1}^{k}||\mathbf{p}_i^\text{sim}-\mathbf{p}_i^\text{gt}||$. However, in our data samples, the length $k$ of the ego trajectory varies. To account for the varied length, the adjusted ADE is used instead: 
    \begin{equation}
        e^{\text{aADE}}=\frac{k_0}{k}e^{\text{ADE}},
    \end{equation}
    where $k_0$ is a fixed pre-selected time step to normalize the error so that the effect of the different trajectory length can be eliminated. This is under the assumption that the error increases in a linear fashion as $k$ increases. 
    
    \item \textit{Adjusted Final Displacement Error (aFDE)}:
    Final displacement error (FDE) is another common metric. Instead of the whole trajectory, it considers the last point on the trajectory: $e^{\text{FDE}}=|\mathbf{p}_k^\text{sim}-\mathbf{p}_k^\text{gt}|$.
    Similar to ADE, this work proposes to use the adjusted FDE:
    \begin{equation}
        e^{\text{aFDE}}=\frac{k_0}{k}e^{\text{FDE}}.
    \end{equation}
    
    \item \textit{Collision Index (CI)}:
    Since the pedestrian model specifically considers vehicle influence, it is necessary to implement a safety-related metric. Therefore, the collision index (CI) is proposed to penalize any collision between the ego pedestrian and the vehicle. $\text{CI}\in [0,1]$ is defined as the proportion of the simulated trajectory that overlaps the vehicle occupancy. $\text{CI}=0$ means no collision.

\end{itemize}

To justify the data-based evaluation, two baseline models are compared with the proposed SG-SFM. The comparison also includes the proposed model with universal calibration and group calibration. These models are summarized as follows:

\begin{itemize}

    \item \textit{Constant Velocity Model}: This model assumes the pedestrian moves toward the final destination $\mathbf{p}_\text{des}$ with the desired speed $v_\text{d}$ without any reaction to other agents. It is denoted as \textit{CV}. 
    
    \item \textit{Ordinary Social Force Model}: The ordinary social force model~\cite{helbing2000simulating} is used as the primary baseline. The friction force is dropped because there is no overcrowded situation, which is a design in the original work for pedestrian evacuation simulation. The original model does not specifically consider vehicle influence, so the vehicle is treated as a rectangular static obstacle composed of the current vehicle occupancy and the predicted vehicle occupancy, as in figure~\ref{fig:rep_veh}. This model is denoted as \textit{SFM}.
    
    \item \textit{Proposed Models}: The proposed models obtained by both universal calibration and group calibration are compared with the above models. They are denoted as \textit{SG-SFM-u} and \textit{SG-SFM-g}, respectively.  
    
\end{itemize}

\subsection{Simulation-based Evaluation}
\label{subse:eval_sim}
Simulation-based evaluation checks the pedestrian motion generated by the process $\mathbb{P}$ for all the scenario configurations $\mathbb{C}\in\mathcal{F}_\mathbb{C}$. $\mathcal{F}_\mathbb{C}$ is the set of fundamental scenarios that are constituted by a series of systematically designed patterns of how the pedestrians are commonly influenced by the vehicle. 

\begin{figure}
    \centering
    \includegraphics[width=\linewidth]{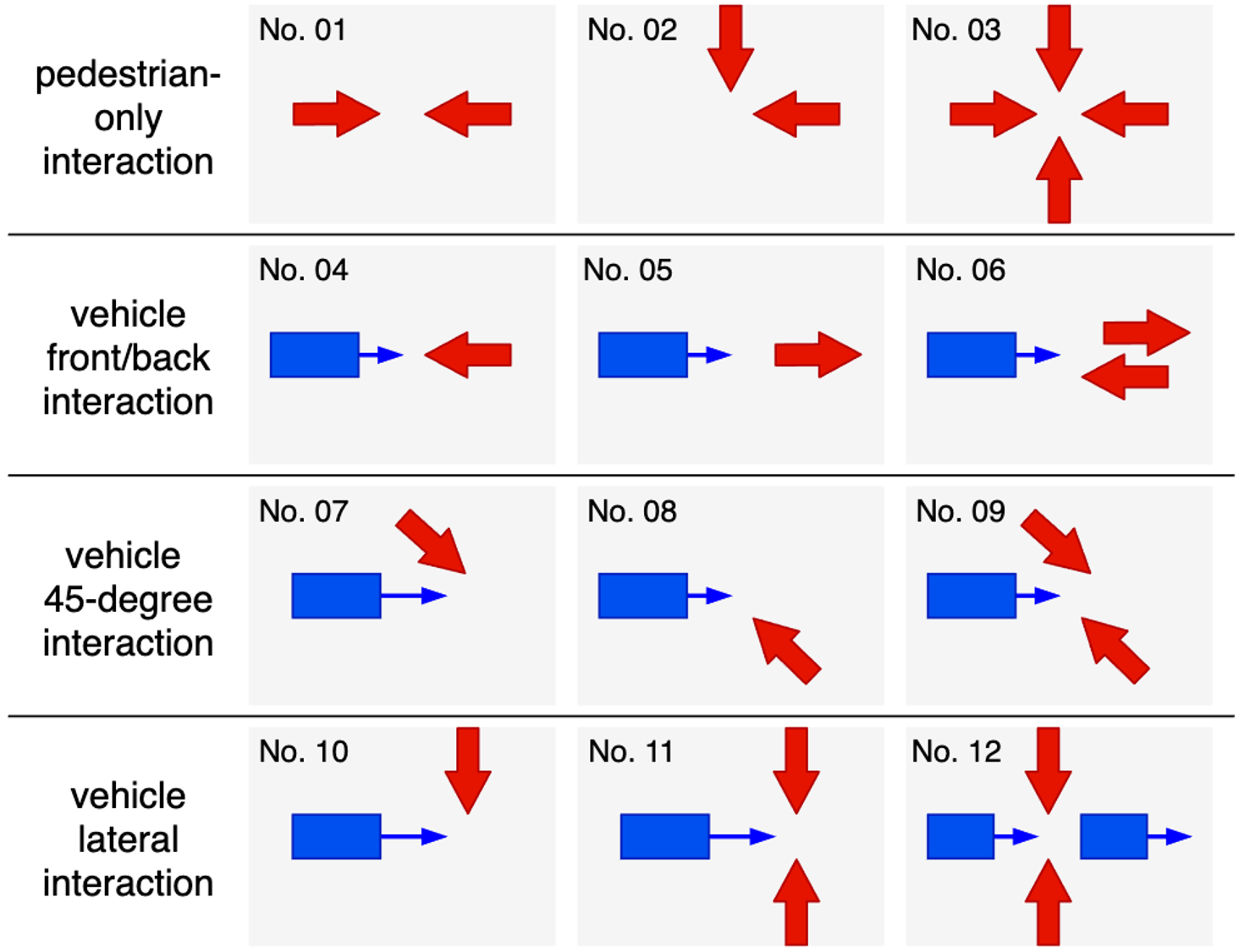}
    \caption{Fundamental scenarios for simulation-based evaluation. Each row represents a category. Blue rectangle with arrow indicates the vehicle and its moving direction. Red arrow indicates the flow of pedestrians. }
    \label{fig:fund_scenarios}
\end{figure}

These scenarios are illustrated in figure~\ref{fig:fund_scenarios}. They are divided into 4 categories (one category in each row): pedestrian-only interaction, vehicle front/back interaction, vehicle 45-degree interaction, and vehicle lateral interaction. For each category, 3 specific scenarios are designed to represent the most common interaction patterns.

In these scenarios, the vehicle motion is designed to maintain a relatively low constant velocity so that it can imitate the cautious driving behavior in shared spaces. The motion is generated using a policy $P_\text{veh}$ that applies a pure-pursuit controller on a bicycle model~~\cite{kuwata2008motion} to follow a pre-defined reference path. The vehicle is not programmed to actively stop for the pedestrians, otherwise in most cases the vehicle just not moves and is to some extent regarded as a static obstacle. More advanced vehicle maneuvers can be easily generated by changing the configurations of $P_\text{veh}$.


\section{Experiments}
\label{se:experiments}

\begin{figure}
    \centering
    \includegraphics[width=0.9\linewidth]{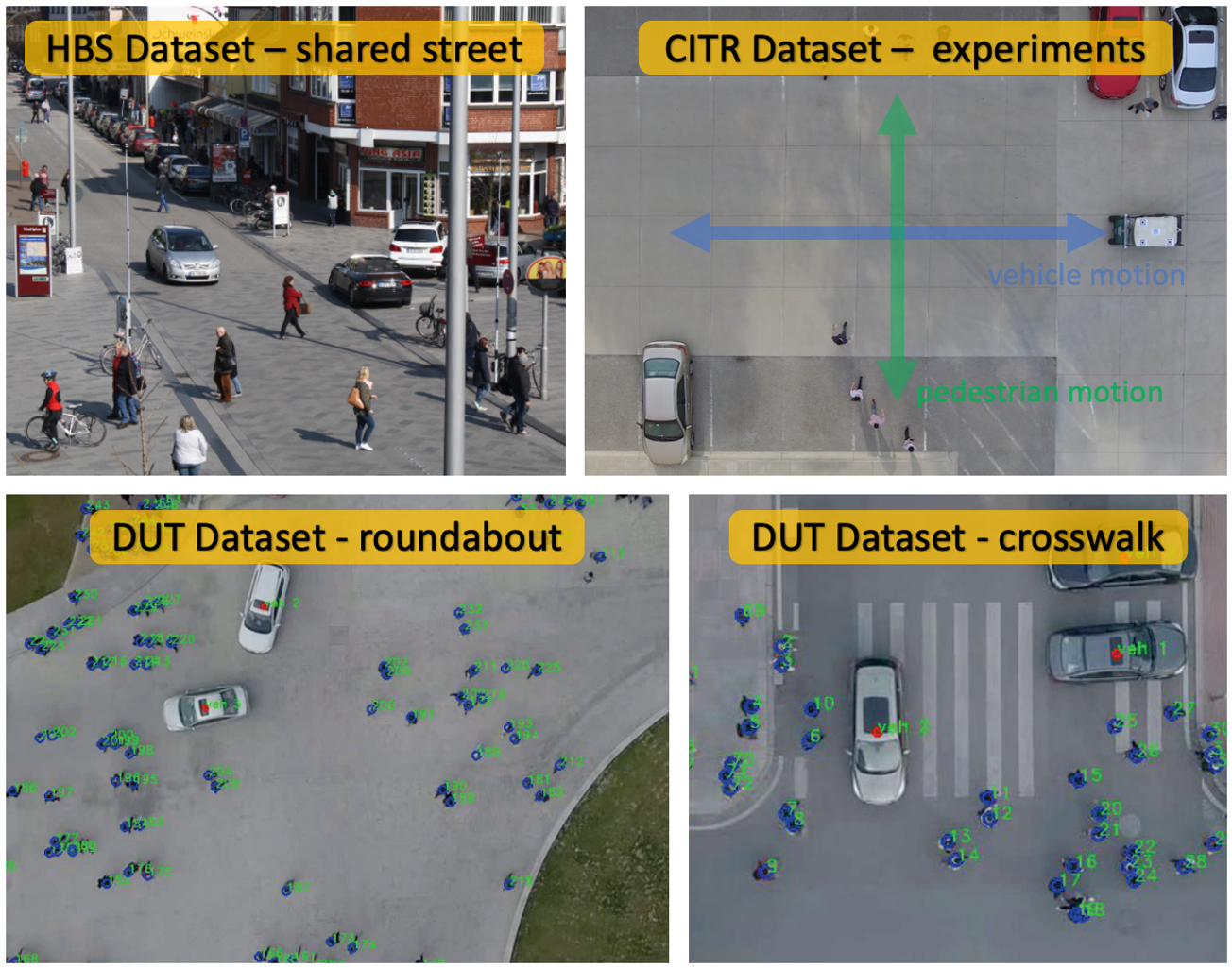}
    \caption{Datasets used for calibration and data-based evaluation.}
    \label{fig:datasets}
\end{figure}

\subsection{Datasets}
\label{subse:dataset}

To verify how well the proposed model generalizes in different real-world scenarios, three different datasets were applied for the parameter calibration and the data-based evaluation. They are Hamburg Bergedorf Station (HBS)~\cite{pascucci2017discrete}, Control and Intelligent Transportation Research Lab~\footnote{Control and Intelligent Transportation Research Lab is associated with the Department of Electrical and Computer Engineering and the Center for Automotive Research at The Ohio State University} (CITR) dataset~\cite{yang2019top}, and Dalian University of Technology (DUT) dataset~\cite{yang2019top}, which primarily cover pedestrian motion under vehicle influence. An illustration of the scenarios in these datasets is shown in figure~\ref{fig:datasets}. HBS is a shared street scenario where low-speed vehicles constantly interact with crossing pedestrians. CITR records controlled experiments where the vehicle interacts with a number of pedestrians from different directions. DUT captures campus scenarios of a shared roundabout and an uncontrolled crosswalk at an intersection. In the process of the data sample preparation, the time step is set as $0.5s$. After the processing, HBS, CITR, and DUT have 199, 208, and 584 data samples, respectively.


\subsection{Implementation}

The experiment was conducted using Python. DEAP framework~\cite{DEAP_JMLR2012} was applied to facilitate the implementation of the genetic algorithm. The genetic algorithm was initialized with a population of 50, with each individual being the tuned parameter values obtained in the previous step. At each iteration, 4 best individuals were kept, the remaining individuals went through the operations of tournament selection, crossover, and mutation. In the clustering step of the group calibration, the number of clusters/groups in K-means was set as 3, which was determined based on the elbow method~\cite{marutho2018determination} over the clustering result from the dataset. 

\begin{table}[]
    \caption{Calibrated Parameters}
    \label{tab:calibrated_params}
    \centering

    \begin{threeparttable}
    
        \begin{tabular}{|l|l|l|l|l|l|l|l|}
            \hline
            \textbf{Type} & $\beta_\text{ped}$ & $\beta_\text{veh}$ & $\tau_\text{x}$ & $d_\text{x}$ & $K_\text{nav}$ & $N_j$ & $d_\text{nav}$ \\ \hline
             
             \hline
            HBS\_universal & 2.99 & 3.60 & 2.00 & 0.50 & 391.06 & 114 & 3.22 \\ \hline
            HBS\_group\_0 & 3.00 & 2.62 & 4.79 & 1.00 & 495.65 & 80 & 6.89 \\ \hline
            HBS\_group\_1 & 3.00 & 3.54 & 2.00 & 0.50 & 800.00 & 94 & 3.00 \\ \hline
            HBS\_group\_2 & 3.00 & 3.57 & 2.00 & 0.50 & 200.00 & 120 & 3.00 \\ \hline
            
            \hline
            CITR\_universal & 3.00 & 3.51 & 2.00 & 0.50 & 286.66 & 86 & 3.74 \\ \hline
            CITR\_group\_0 & 2.97 & 3.60 & 2.04 & 0.51 & 247.91 & 82 & 3.41 \\ \hline
            CITR\_group\_1 & 3.00 & 3.58 & 2.00 & 0.50 & 271.75 & 80 & 3.00 \\ \hline
            CITR\_group\_2 & 3.00 & 3.25 & 2.09 & 0.50 & 324.49 & 80 & 5.23 \\ \hline
            
            \hline
            DUT\_universal & 3.00 & 3.60 & 2.00 & 0.50 & 237.98 & 80 & 3.00 \\ \hline
            DUT\_group\_0 & 2.98 & 3.53 & 2.00 & 0.50 & 200.00 & 80 & 3.00 \\ \hline
            DUT\_group\_1 & 3.00 & 3.26 & 2.01 & 0.50 & 243.09 & 102 & 3.00 \\ \hline
            DUT\_group\_2 & 3.00 & 3.60 & 2.00 & 0.68 & 238.74 & 80 & 3.00 \\ \hline
        \end{tabular}
        
    \end{threeparttable}

\end{table}

In the data-based evaluation, the normalizing time step is set to $k_0=10$, which corresponds to a simulated horizon of $5s$ with the time step $\Delta t=0.5s$. This selection roughly matches the commonly applied evaluation horizon of $4.8s$ in pedestrian-only evaluation~\cite{pellegrini2009you, alahi2017learning}, which evaluates $12$ frames with a different time step of $0.4s$ due to the data annotation. In the simulation-based evaluation, each scenario in figure~\ref{fig:fund_scenarios} was simulated using different number $n_\text{ped}^\text{sim}$ of pedestrians in each pedestrian flow (red arrow in figure~\ref{fig:fund_scenarios}). A combination of $n_\text{ped}^\text{sim}=1,5,10$ was determined to cover different pedestrian densities. The vehicle motion is set to cruise at a speed of $2.0m/s$.



\section{Results}
\label{se:results}
\subsection{Data-based Evaluation Results}

\begin{table}

    \caption{Data-based Evaluation Results\tnote{*}}
    \label{tab:scores}
    
    \centering
    
    \begin{threeparttable}
    
        \begin{tabular}{|l|l|l|l|}
        \hline
            \textbf{Method} & \textbf{HBS Dataset} & \textbf{CITR Dataset} & \textbf{DUT Dataset}\\ \hline
            
            \hline
            CV & 0.897/1.140/0.023 & 0.378/0.481/0.020 & 0.306/0.293/0.026 \\ \hline
                    
            \hline
            SFM & 0.896/1.111/0.009 & 0.455/0.711/0.003 & \textbf{0.544}/\textbf{0.857}/0.009 \\ \hline
            
            SG-SFM-u & \textcolor[rgb]{0,0,1}{0.848}/\textcolor[rgb]{0,0,1}{0.976}/\textbf{0.002} & \textcolor[rgb]{0,0,1}{0.408}/\textcolor[rgb]{0,0,1}{0.627}/\textbf{0.001} & 0.597/0.978/\textbf{0.005} \\ \hline
            
            SG-SFM-g & \textbf{0.777}/\textbf{0.967}/\textbf{0.002} & \textbf{0.392}/\textbf{0.592}/\textbf{0.001} & \textcolor[rgb]{0,0,1}{0.589}/\textcolor[rgb]{0,0,1}{0.958}/\textbf{0.005} \\ \hline
        \end{tabular}
    
        \begin{tablenotes}
            \footnotesize
            \item[*] Each entry reports the results of aADE/aFDE/CI. 
            \item[**]  The {\bf bold} number indicates the best among the last three methods, and the {\textcolor[rgb]{0,0,1}{blue}} indicates the second best.
        \end{tablenotes}
    
    \end{threeparttable}
\end{table}

\begin{figure}
    \centering
    \includegraphics[width=\linewidth]{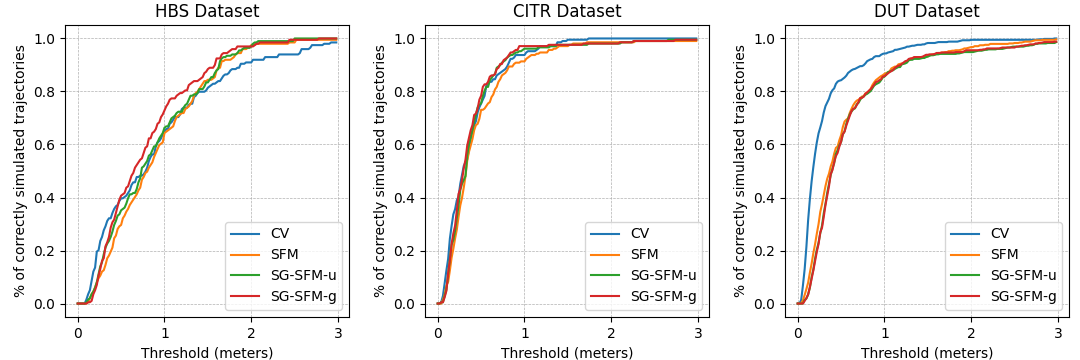}
    \caption{Threshold of aADE vs. the percentage correctly simulated trajectory.}
    \label{fig:thr_vs_correct}
\end{figure}

\begin{figure*}
    \centering
    \includegraphics[width=\linewidth]{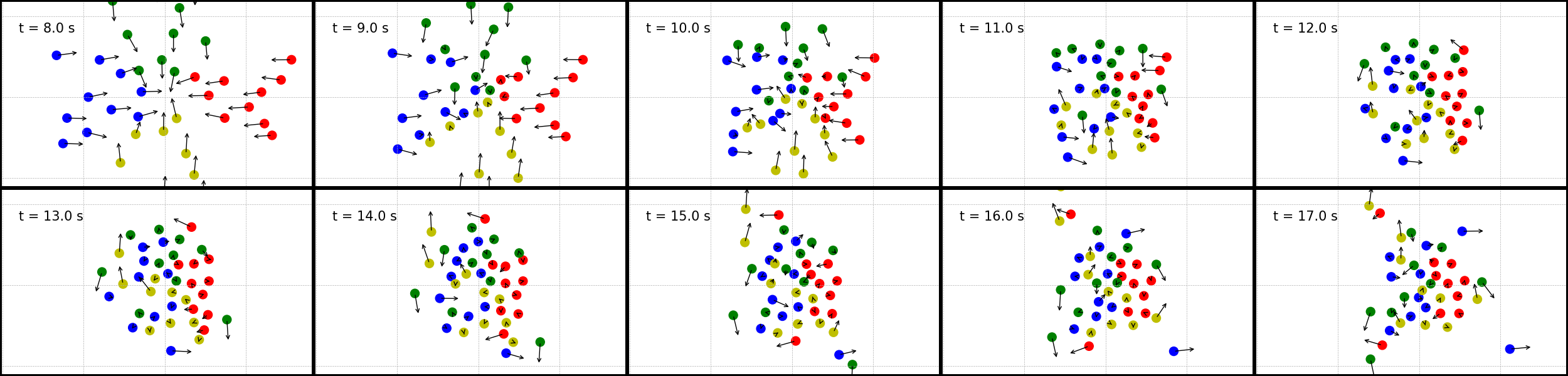}
    \caption{Simulation snapshots of the last scenario of the category of pedestrian-only interaction (row 1 and column 3 in figure~\ref{fig:fund_scenarios}). Pedestrians (small circles) belonging to different flows are marked by different colors. The arrow associated with each pedestrian indicates the velocity vector.}
    \label{fig:sim_cat_1}
\end{figure*}

\begin{figure*}
    \centering
    \includegraphics[width=\linewidth]{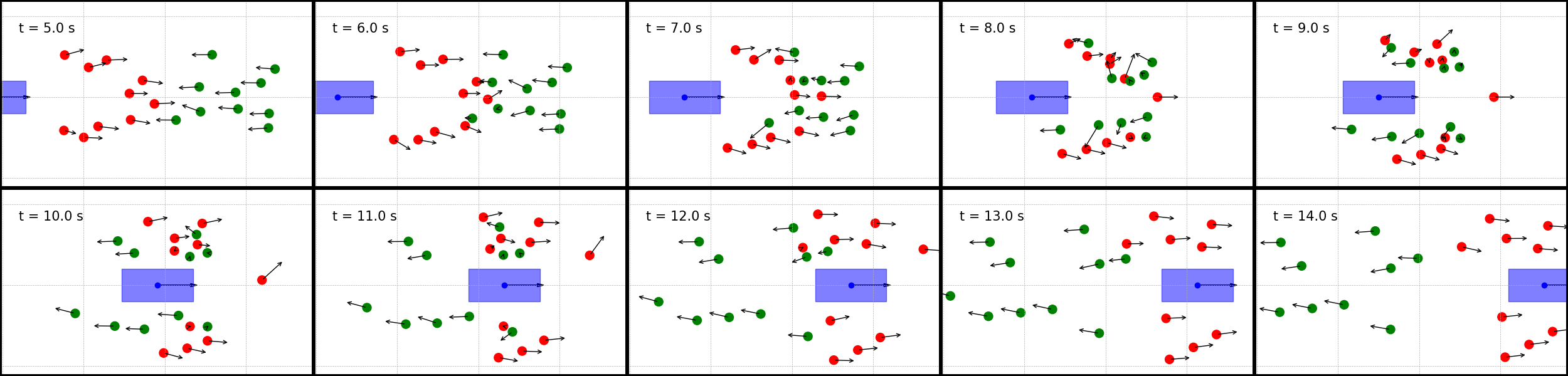}
    \caption{Simulation snapshots of the last scenario of the category of vehicle front/back interaction (row 2 and column 3 in figure~\ref{fig:fund_scenarios}). Pedestrians (small circles) belonging to different flows are marked by different colors. Blue rectangle is the vehicle. The arrow indicates the velocity vector.}
    \label{fig:sim_cat_2}
\end{figure*}

\begin{figure*}
    \centering
    \includegraphics[width=\linewidth]{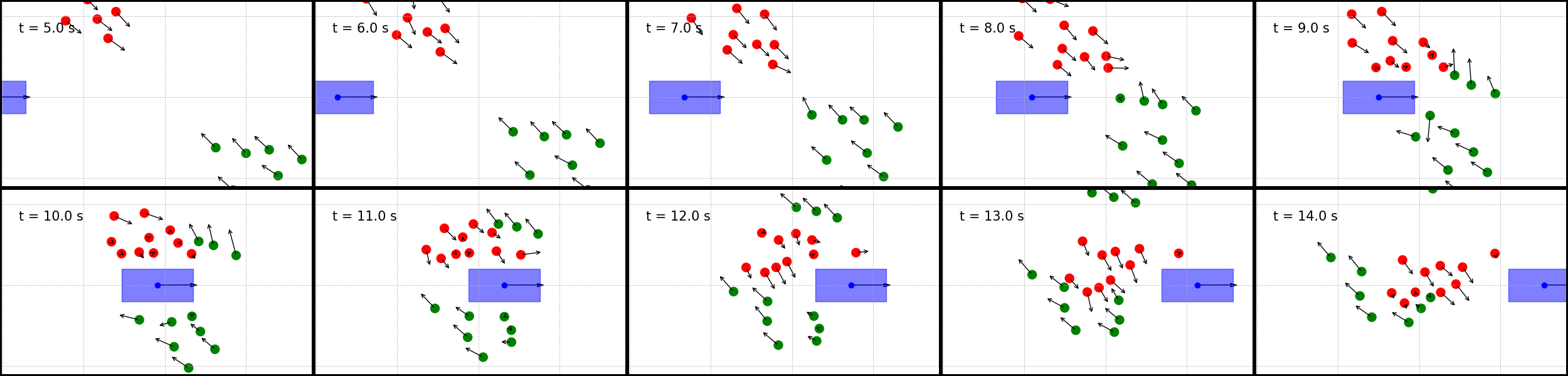}
    \caption{Simulation snapshots of the last scenario of the category of vehicle 45-degree interaction (row 3 and column 3 in figure~\ref{fig:fund_scenarios}). Pedestrians (small circles) belonging to different flows are marked by different colors. Blue rectangle is the vehicle. The arrow indicates the velocity vector.}
    \label{fig:sim_cat_3}
\end{figure*}

\begin{figure*}
    \centering
    \includegraphics[width=\linewidth]{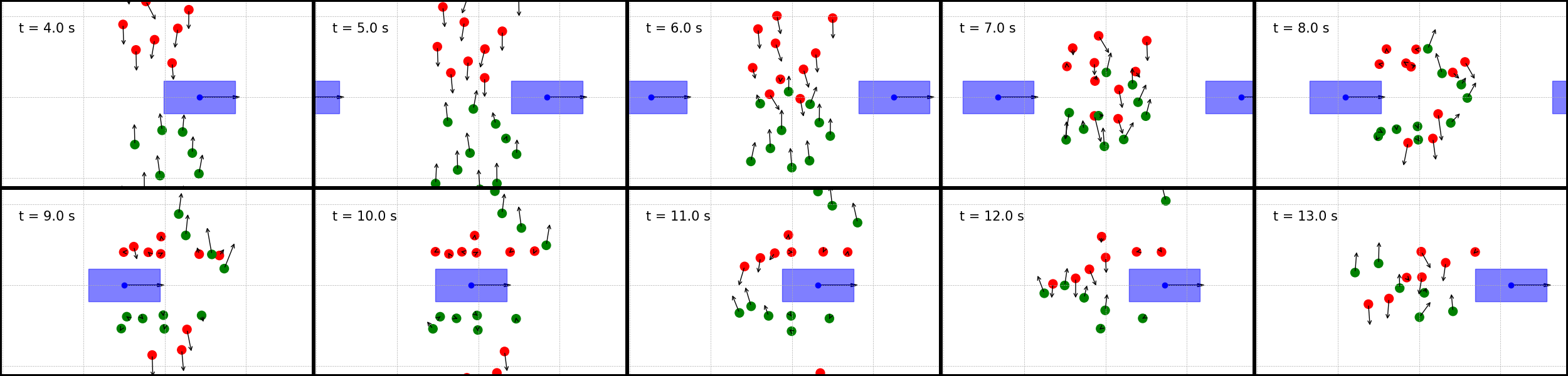}
    \caption{Simulation snapshots of the last scenario of the category of vehicle lateral interaction (row 4 and column 3 in figure~\ref{fig:fund_scenarios}). Pedestrians (small circles) belonging to different flows are marked by different colors. Blue rectangle is the vehicle. The arrow indicates the velocity vector.}
    \label{fig:sim_cat_4}
\end{figure*}

Table~\ref{tab:calibrated_params} displays the calibrated parameters for each dataset and each mode of calibration. Table~\ref{tab:scores} shows the data-based evaluation results. In each entry of table~\ref{tab:scores}, the reported 3 scores are aADE, aFDE, and CI, respectively. The CV baseline has better aADE and aFDE scores in CITR and DUT datasets. However, CV baseline generates the highest CI scores for all datasets, which is unacceptable. Therefore, the following analysis is based on the comparison among SFM, SG-SFM-u, and SG-SFM-g, excluding CV. 

For each dataset, the bold number indicates the best score, and the blue number indicates the second-best score. Overall, the proposed SG-SFM achieved better performance. In HBS and CITR datasets, SFM is outperformed by SG-SFM-u/SG-SFM-g in all metrics. In the DUT dataset, SFM is slightly better than SG-SFM-u/SG-SFM-g in aADE and aFDE, but not in CI. This is acceptable because based on the design of SFM, there is a trade-off between the aADE/aFDE score and the CI. In SFM, the CI can be reduced by tuning the parameters at the cost of increased aADF/aFDE. 

Since existing works do not have a quantitative result in our configuration, we need to verify that these metric scores fall into a reasonable range. Our aADE/aFDE scores are comparable to the state-of-the-art ADE/FDE scores for the models that only consider pedestrian-to-pedestrian interaction~\cite{sadeghian2019sophie}. Their ADE/FDE scores are around the values of $0.5m/1.0m$ that account for the motion of $4.8s$. In our work, with additional consideration of vehicle influence, we still get comparable results, which validates the quantitative scores.

Comparing SG-SFM-u with SG-SFM-g, SG-SFM-g is always better than SG-SFM-u. This is consistent with what we expected. Introducing different characteristics for the pedestrian motion essentially increases the model complexity. As a result, SG-SFM-g should have better performance.

Figure~\ref{fig:thr_vs_correct} shows the threshold of aADE versus the percentage of correctly simulated trajectory. For a particular threshold $T$ (a value in x axis), if the $e^\text{aADE}<T$, then the corresponding trajectory is considered to be correctly simulated. The results of these graphs are in accordance with the results in table~\ref{tab:scores}. For HBS and CITR datasets, SG-SFM-g is the best for most range of the threshold, while SG-SFM-u is the second best. In the DUT dataset, both SG-SFM-g and SG-SFM-u achieve comparable results to SFM. Note that the CV baseline is much better than other methods in the DUT dataset. A possible reason is that the DUT dataset contains more linear motion trajectories than the other two datasets.

\subsection{Simulation of Fundamental Scenarios}

\begin{figure}
    \centering
    \includegraphics[width=\linewidth]{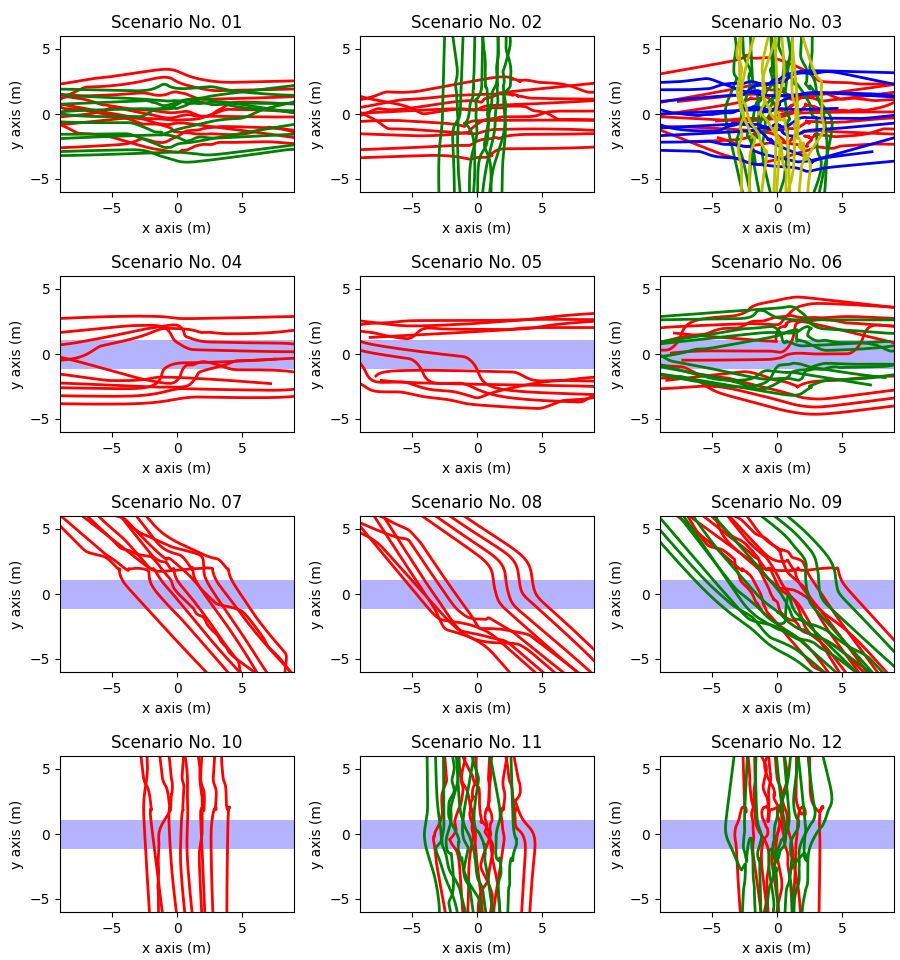}
    \caption{Pedestrian trajectories in the simulation of different fundamental scenarios. Each small figure corresponds to a scenario in the same position of figure~\ref{fig:fund_scenarios}. Different pedestrian flows are differentiated by different colors. The blue stripe shows the trace of the vehicle. }
    \label{fig:sim_trajs}
\end{figure}

Figure~\ref{fig:sim_trajs} shows the simulated pedestrian trajectories in all fundamental scenarios. The position of each scenario in figure~\ref{fig:sim_trajs} matches the position in figure~\ref{fig:fund_scenarios}. Here the simulated trajectories are generated using the model calibrated over the DUT dataset using universal calibration. The simulations of the models calibrated with other configurations have similar performance. In the following descriptions, the simulation results are analyzed category by category (row by row). For each category, the simulation snapshots of the last scenario (last column in figure~\ref{fig:fund_scenarios}) are also created to investigate the detailed pedestrian motion. The last scenarios are chosen because they cover all the interaction patterns in the corresponding categories. The snapshots are shown in figure~\ref{fig:sim_cat_1}, figure~\ref{fig:sim_cat_2}, figure~\ref{fig:sim_cat_3}, and figure~\ref{fig:sim_cat_4}, respectively. In these figures, a blue rectangle indicates a vehicle, while a small circle indicates a pedestrian. The associated arrow on a pedestrian or a vehicle indicates the velocity vector. Different pedestrian flows are differentiated by different colors. 

In the first category of pedestrian-only interaction (1st row in figure~\ref{fig:sim_trajs}), the pedestrians are able to navigate around and avoid collision with surrounding pedestrians. This is further demonstrated in figure~\ref{fig:sim_cat_1} by showing the snapshots of scenario No. 3. Also, by observing the simulated trajectories of scenario No. 1 in figure~\ref{fig:sim_trajs}, when two pedestrian flows encounter each other, the lane formation phenomenon can be identified. This is a common observation of reasonable pedestrian behavior~\cite{helbing1995social} in pedestrian-to-pedestrian interaction.  

In the second category of vehicle front/back interaction (2nd row in figure~\ref{fig:sim_trajs}), the pedestrians successfully create space for the vehicle in both the front and back interaction, which indicates the pedestrians can properly navigate around the incoming vehicle. Specifically, figure~\ref{fig:sim_cat_2} shows the snapshots of scenario No. 6, we can further confirm that there is no collision between pedestrians and the interacting vehicle.

In the third category of vehicle 45-degree interaction (3rd row in figure~\ref{fig:sim_trajs}), the pedestrians show interesting behaviors. Taking the snapshots of scenario No. 9, as illustrated in figure~\ref{fig:sim_cat_3}, for example, when the vehicle is still far enough, some pedestrians decide not to yield and continue to move. While the vehicle is approaching, they accelerate a little bit to clear the danger. In contrast, some other pedestrians decide to yield because the vehicle is quite close. In the process of yielding, they slowly move in the direction parallel to the vehicle moving direction so that they can reach the destination sooner. The above behaviors are exactly the phenomena commonly observed in real-world situations.

In the last category of vehicle lateral interaction (4th row in figure~\ref{fig:sim_trajs}), in addition to successfully avoiding the collision with the vehicle, it is also observed that some pedestrians are trying to find a gap between the two vehicles. For example in scenario No. 12, as illustrated by the snapshots in figure~\ref{fig:sim_cat_4}. Some pedestrians decide to continue moving in between the two vehicles when the later vehicle is still at a certain distance away from the interaction. Other pedestrians decide to wait until the later vehicle has passed in interaction area. In the whole process, there is no collision with vehicles. This is also a common behavior in real-world situations.

In sum, the above observation and analysis in simulation-based evaluation demonstrated that the proposed SG-SFM can generate realistic pedestrian behavior under different vehicle-pedestrian interaction scenarios. Together with the data-based evaluation, the effectiveness of the proposed model was successfully verified.

\section{Conclusion}
\label{se:conclusion}

This work proposed the sub-goal social force model (SG-SFM) for the collective pedestrian motion under vehicle influence. The proposed model builds on the ordinary social force model with the integration of the sub-goal concept so that the influence from the surrounding pedestrians and vehicles can be easily combined. The model was successfully calibrated in two different ways over three different datasets. The quantitative evaluation over the three different datasets demonstrated the model generalization in different scenarios. Furthermore, a simulation-based evaluation over a series fundamental scenarios of different types of vehicle influences further demonstrated the effectiveness of the model.

Future improvements include the model design itself and the process of the calibration and the evaluation. First, the modeling of the vehicle influence can be improved. The major challenge of generic pedestrian motion modeling lies in the coverage of unusual vehicle-pedestrian interaction patterns. 
To address this issue, data-driven designs are worth exploring, as the dataset may contain some interaction patterns that are unnoticed by designers. This further requires the collection and creation of more datasets of various scenarios. Second, there could be a better way of calibrating the model, depending on how the model is designed. If the model is designed in such a way that some gradient-based calibration approach is applicable, it will be a huge progress in this field. Last, some quantitative criteria such as evaluating pedestrian's local density or squeezing phenomena can be designed to justify the pedestrian motion in the simulation-based evaluation. However, properly designing such criteria requires further investigation. Besides that, fundamental scenarios can also be extended to cover more vehicle maneuvers.









%

\appendices


\section*{Acknowledgment}


This work was supported by the United States Department of Transportation under (\#69A3551747111) for the Mobility21 University Transportation Center and by the German Research Foundation (DFG) through the Research Training Group SocialCars (GRK 1931). We acknowledge the DFG research project MODIS (\#248905318) for sharing the HBS dataset.

\ifCLASSOPTIONcaptionsoff
  \newpage
\fi



\bibliographystyle{IEEEtran}
\bibliography{mybib}

\begin{IEEEbiography}[{\includegraphics[width=1in,height=1.25in,clip,keepaspectratio]{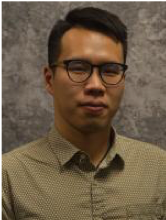}}]{Dongfang Yang}

Dongfang Yang received his bachelor's degree in microelectronics from Sun Yat-sen University, Guangzhou, China, in 2014. He has been with The Ohio State University since 2015 and received his Ph.D. in Electrical and Computer Engineering from The Ohio State University, in 2020. He is currently a graduate research associate at The Ohio State University. His research interests include control systems, computer vision, and machine learning with applications in intelligent transportation and autonomous driving.

\end{IEEEbiography}

\begin{IEEEbiography}[{\includegraphics[width=1in,height=1.25in,clip,keepaspectratio]{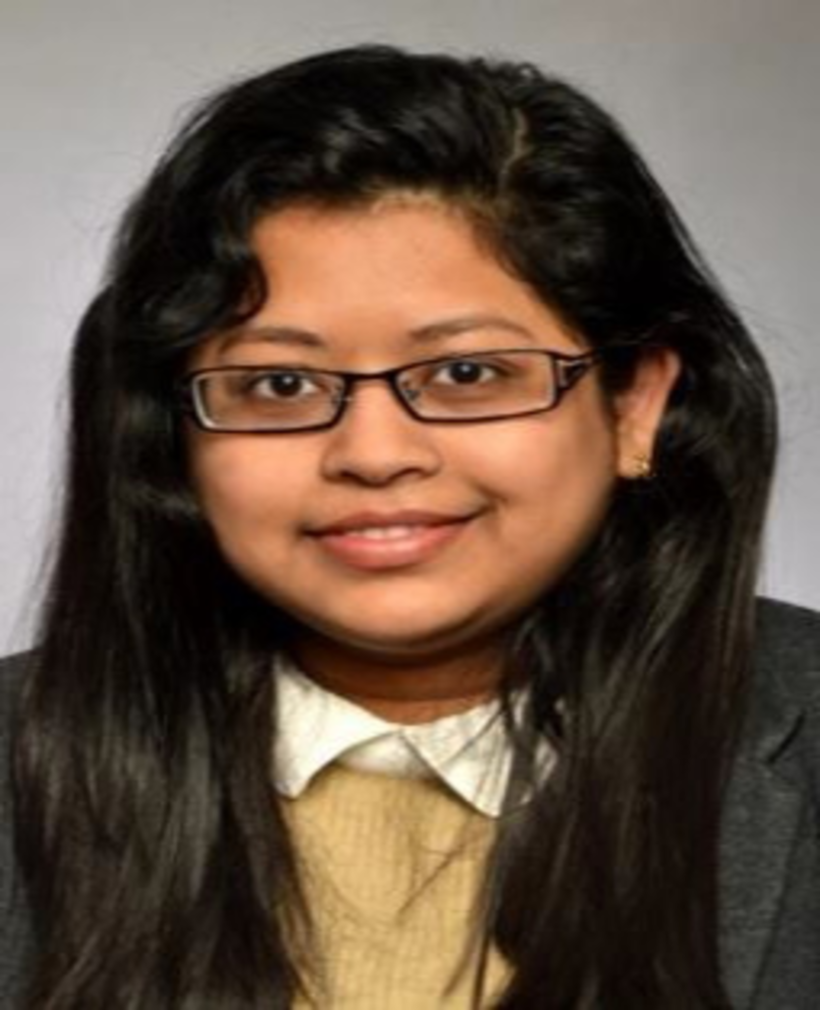}}]{Fatema T. Johora} received her B.Sc. degree in Computer Science and Engineering from Jessore University of Science and Technology, Bangladesh, in 2013, and her M.Sc. degree in Internet Technologies and Information Systems from Clausthal University of Technology, Germany, in 2017. She is currently a doctoral candidate at the Department of Informatics at Clausthal University of Technology. 

Her research interests cover game theory, agent-based modeling, and machine learning in the area of intelligent transport system and autonomous driving.
\end{IEEEbiography}

\begin{IEEEbiography}[{\includegraphics[width=1in,height=1.25in,clip,keepaspectratio]{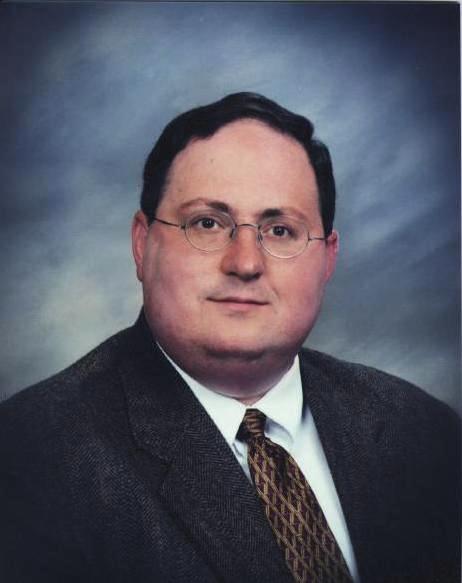}}]{Keith A. Redmill}
(S’89–M’98–SM’11) received the B.S.E.E. and B.A. degrees in mathematics from Duke University, Durham, NC, USA, in 1989 and the M.S. and Ph.D. degrees from The Ohio State University, Columbus, OH, USA, in 1991 and 1998, respectively. Since 1998, he has been with the Department of Electrical and Computer Engineering, The Ohio State University, initially as a Research Scientist. He is currently a Research Associate Professor.

He is a coauthor of the book \textit{Autonomous Ground Vehicles}. His research interests include autonomous vehicles and robots, intelligent transportation systems, vehicle and bus tracking, wireless data communication, cellular digital packet data, Global Positioning System and Geographic Information System technologies, large hierarchical systems, real-time and embedded systems, hybrid systems, control theory, dynamical systems theory, cognitive science, numerical analysis and scientific computation, and computer engineering.
\end{IEEEbiography}

\begin{IEEEbiography}[{\includegraphics[width=1in,height=1.25in,clip,keepaspectratio]{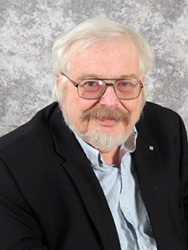}}]{Ümit Özgüner} (S’72–M’75–F’10) Prof. Emeritus Ümit Özgüner, TRC Inc. Chair on ITS at The Ohio State University, is a well know expert on Intelligent Vehicles. He holds the title of “Fellow” in IEEE for his contributions to the theory and practice of autonomous ground vehicles and is the Editor in Chief of the IEEE ITS Society, Transactions on Intelligent Vehicles.

He has led and participated in many autonomous ground vehicle related programs like DoT FHWA Demo’97, DARPA Grand Challenges and the DARPA Urban Challenge. His research has been (and is) supported by many industries including Ford, GM, Honda and Renault. He has published extensively on control design and vehicle autonomy and has co-authored a book on Ground Vehicle Autonomy. His present projects are on Machine Learning for driving, pedestrian modeling at OSU and participates externally on V\&V and risk mitigation, and self-driving operation of specialized vehicles. Professor Ozguner has developed and taught a course on Ground Vehicle Autonomy for many years and has advised over 35 students during their studies towards a PhD.
\end{IEEEbiography}

\begin{IEEEbiography}[{\includegraphics[width=1in,height=1.25in,clip,keepaspectratio]{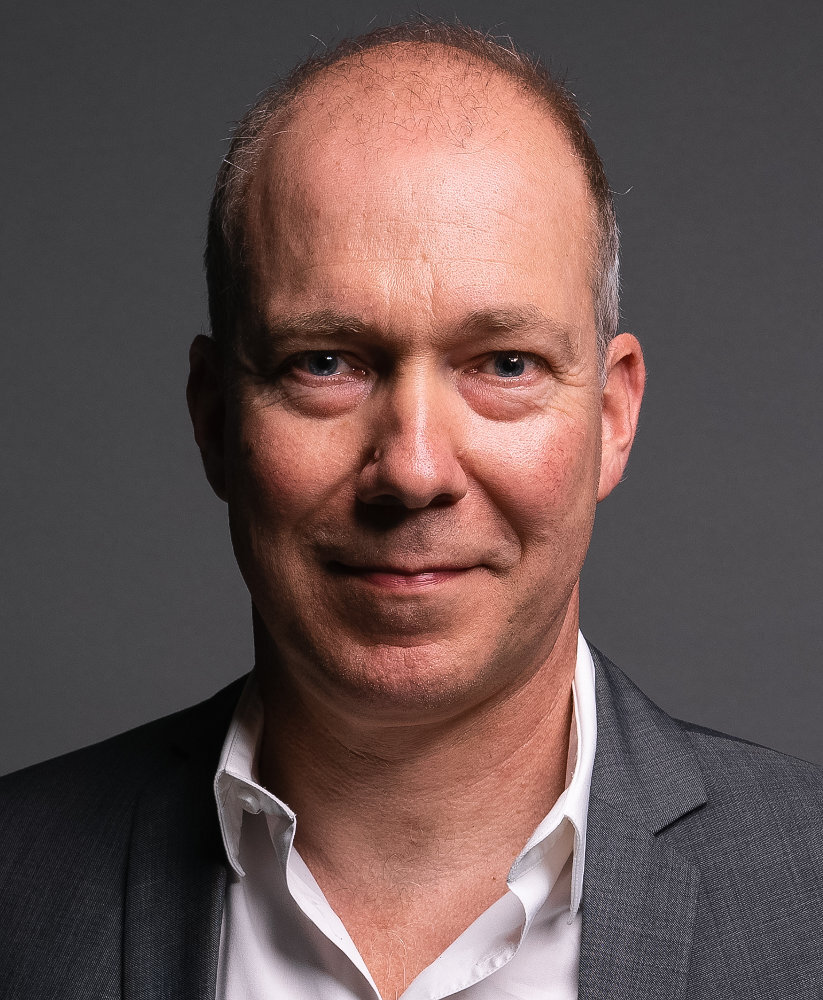}}]{Jörg P. Müller} is a Full Professor of Computer Science at the Department of Informatics at Clausthal University of Technology. He holds a Ph.D. in Computer Science from Universität des Saarlandes in the area of Artificial Intelligence, Intelligent Agents and Multiagent Systems. Prior to becoming a professor, he obtained ten years of industrial research experience in agent technology and peer-to-peer computing, working for Mitsubishi Electric, John Wiley and Sons, and Siemens Corporate Technology. His current research interests cover the broad area of modelling and simulation of socio-technical systems, coordination and intelligent systems. A long-term research focus is on agent-based modelling and simulation in the area of intelligent transport systems and future connected traffic systems.  Jörg has served on numerous conference committees in the area of AI, intelligent agents and multi-agent systems and has co-authored over 250 scientific publications.
\end{IEEEbiography}

\vfill
\vfill

\end{document}